\documentclass[10pt,twocolumn,letterpaper]{article}

\usepackage{iccv}
\usepackage{times}
\usepackage{epsfig}
\usepackage{graphicx}
\usepackage{amsmath}
\usepackage{amssymb}
\usepackage{booktabs}
\usepackage{xcolor}
\usepackage{colortbl}
\usepackage{multirow}
\usepackage{multicol}
\usepackage{subfigure}
\usepackage{amsthm}

\usepackage[pagebackref=true,breaklinks=true,letterpaper=true,colorlinks,bookmarks=false]{hyperref}

\iccvfinalcopy 

\usepackage[capitalize]{cleveref}
\crefname{section}{Sec.}{Secs.}
\Crefname{section}{Section}{Sections}
\Crefname{table}{Table}{Tables}
\crefname{table}{Tab.}{Tabs.}

\def\iccvPaperID{6040} 

\ificcvfinal\fi

\begin{document}

\title{FLatten Transformer: Vision Transformer using Focused Linear Attention}

\author{Dongchen Han\thanks{Equal contribution.}~~~ Xuran Pan$^{*}$~~~Yizeng Han~~~Shiji Song~~~Gao Huang$\thanks{Corresponding Author.}$ \\ 
{\small Department of Automation, BNRist, Tsinghua University}\\
}



\maketitle
\ificcvfinal\thispagestyle{empty}\fi


\begin{abstract}

The quadratic computation complexity of self-attention has been a persistent challenge when applying Transformer models to vision tasks. Linear attention, on the other hand, offers a much more efficient alternative with its linear complexity by approximating the Softmax operation through carefully designed mapping functions. However, current linear attention approaches either suffer from significant performance degradation or introduce additional computation overhead from the mapping functions. In this paper, we propose a novel \textbf{Focused Linear Attention} module to achieve both high efficiency and expressiveness. Specifically, we first analyze the factors contributing to the performance degradation of linear attention from two perspectives: the \textit{focus ability} and \textit{feature diversity}. To overcome these limitations, we introduce a simple yet effective mapping function and an efficient rank restoration module to enhance the expressiveness of self-attention while maintaining low computation complexity. Extensive experiments show that our linear attention module is applicable to a variety of advanced vision Transformers, and achieves consistently improved performances on multiple benchmarks. Code is available at \url{https://github.com/LeapLabTHU/FLatten-Transformer}.

\end{abstract}

\section{Introduction} \label{sec:introduction}

\begin{figure}[t]
    \centering
    \includegraphics[width=\linewidth]{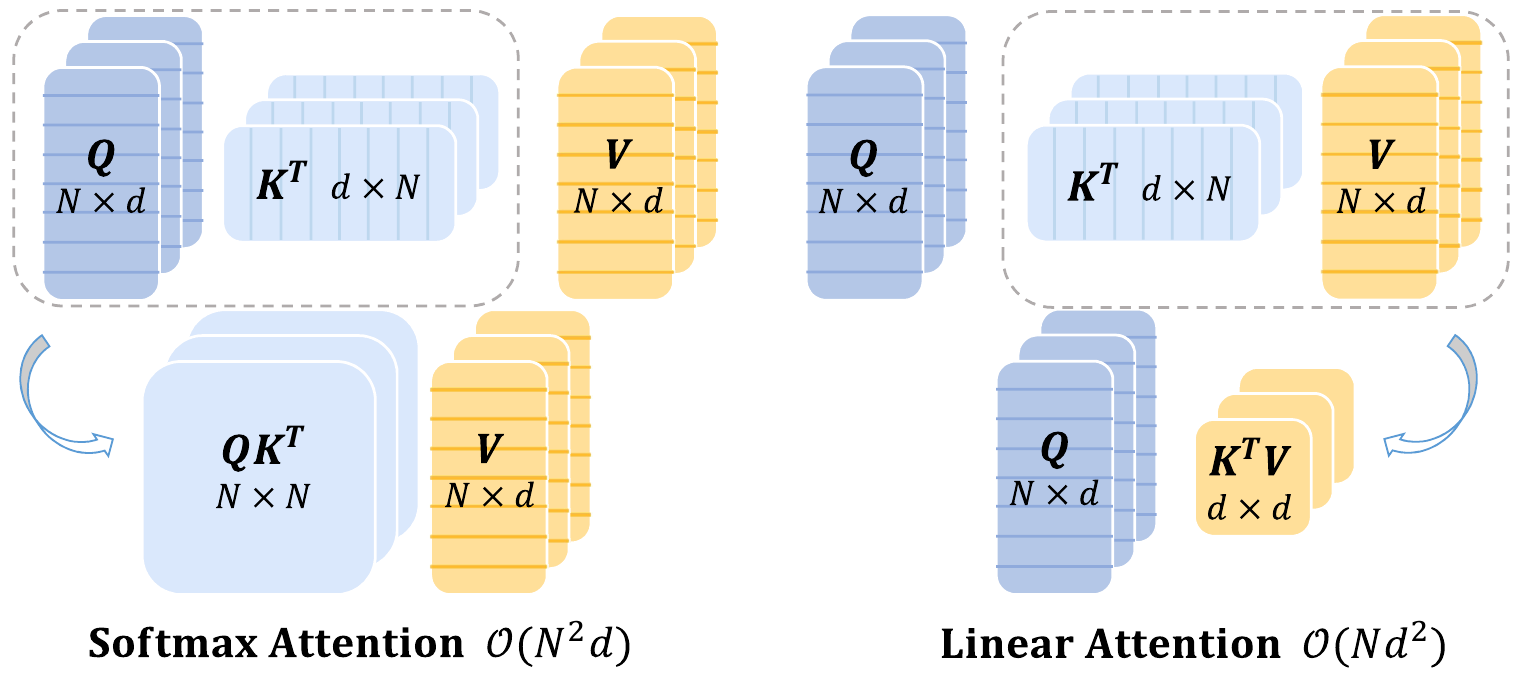}
    \caption{\textbf{Difference between Softmax attention and Linear attention.} $Q, K, V \!\in\! \mathbb{R}^{N\!\times\!d}$ denote query, key and value matrix respectively.
    Softmax attention compels to compute the pairwise similarity between queries and keys, and results in the complexity of $\mathcal{O}(N^2d)$. Linear attention manages to decouple the Softmax operation with proper approximation and change the computation order by computing $K^TV$ first, which leads to the complexity of $\mathcal{O}(Nd^2)$. Considering that channel dimension $d$ is usually smaller than token number $N$ in modern vision Transformer designs, \textit{e.g.,} $d\!=\!64,N\!=\!196$ in DeiT~\cite{deit} and $d\!=\!32,N\!=\!49$ in Swin Transformer~\cite{swin}, linear attention modules practically save the overall computation cost while can also enjoy the benefits of a larger receptive field and higher throughput.}
    \label{fig:softmax_linear_attn}
    \vskip -0.05in
\end{figure}

Recent years have witnessed the vast development of Transformer and self-attention in the field of computer vision. With the advent of Vision Transformer~\cite{vit,deit}, self-attention techniques have shown great potential in a variety of vision tasks including image classification~\cite{pvt,dvt,acmix,dat}, semantic segmentation~\cite{maskformer,segformer}, object detection~\cite{detr,deformabledetr,vitdet}, and multi-modal tasks~\cite{clip,knowledge_clip}.

However, applying Transformer to vision models is a non-trivial task. Unlike lightweight convolution neural networks~\cite{mobilenetv2,lasnet,gfnet,rotated_conv}, the quadratic computation complexity $\mathcal{O}(n^2)$ with respect to sequence length $n$ leads to high computation costs when employing self-attention with a global receptive field. Previous works have sought to mitigate this challenge by confining the global receptive field to a smaller region, such as designing sparse global attention patterns~\cite{pvt,dat} or applying smaller attention windows~\cite{swin,nat}. 
Albeit effective, these methods are either prone to disregarding informative features in other regions due to their attention patterns or inevitably sacrifice the ability to model long-range dependencies.

Linear attention, on the other hand, has been considered a simple yet effective alternative to address the computation dilemma by reducing the general complexity. Early research leverages a locally-sensitive hashing scheme~\cite{reformer} that compresses the computation complexity from $\mathcal{O}(n^2)$ to $\mathcal{O}(n{\rm log}(n))$. Nevertheless, it introduces a large constant before the complexity term, which makes it still unaffordable under common cases. More recent studies have noticed that the utilization of Softmax function in the self-attention operation practically compels a pairwise computation between all queries and keys, resulting in the predominant $\mathcal{O}(n^2)$ complexity. To tackle this, several approaches adopt simple activation functions~\cite{linear_attn,efficient_attn} or tailored mapping functions~\cite{performer,soft} to approximate the original Softmax function. As illustrated in Fig.~\ref{fig:softmax_linear_attn}, by changing the computation order from (query·key)·value to query·(key·value), the overall computation complexity can be reduced to $\mathcal{O}(n)$. However, compared to Softmax attention, current linear attention approaches still suffer from severe performance drop and may involve additional computation overhead from the mapping function, thereby constraining their practical application.

In this paper, we target on the limitations of current linear attention approaches and propose a novel \textbf{Focused Linear Attention} module, which achieves both high efficiency and expressiveness. Specifically, we undertake a dual-pronged analysis of the factors contributing to the performance decline in linear attention and subsequently propose corresponding solutions. First, the distribution of attention weight in the former linear attention modules is relatively smooth, lacking the focus ability to address the most informative features. As a remedy, we propose a simple mapping function to adjust the feature direction of queries and keys, making the attention weights more distinguishable. Second, we notice that the diminished rank of the attention matrix curtails the diversity of features in linear attention. To address this, we propose a rank restoration module by applying an additional depthwise convolution (DWC) to the original attention matrix, which helps to restore the matrix rank and keeps the output feature of different positions diversified. Leveraging these improved techniques, our module demonstrates comparable or superior performance to its Softmax counterparts, while enjoying the benefits of low computation complexity.

We empirically validate the effectiveness of our module on image classification, semantic segmentation, and object detection tasks using five advanced vision Transformer models. The results demonstrate consistent improvements over all baselines and other linear attention approaches.

\section{Related Works} \label{sec:related_works}
\subsection{Vision Transformer}
Transformer and self-attention mechanism are first introduced in the field of natural language processing and have earned wide research interest in computer vision. Nevertheless, the high computation complexity of self-attention set constraints on the direct application to vision tasks. Previous works have attempted to address this concern from several perspectives. The pioneer Vision Transformer~\cite{vit} considers reducing the input resolution by merging neighbouring pixels into a single token. Similar insights have been adopted in the following researches~\cite{volo,t2tvit} and also extend to downstream tasks~\cite{vitdet}. 
Another line of research reduces the feature resolution gradually and adopts carefully designed attention patterns to constrain the number of attentive tokens. 
For instance, PVT~\cite{pvt,pvtv2} uses a sparse attention pattern and selects attentive tokens from a global perspective. DAT~\cite{dat} follows the path and designs a deformable attention module to achieve data-dependent attention pattern. Swin Transformer~\cite{swin} selects attentive tokens locally by dividing input into isolated windows. NAT~\cite{nat} follows the query-centric pattern in convolution and designs independent attentive tokens for all queries. Some researches also notice that convolution operations are valuable to Transformer models and may help to improve the overall efficiency~\cite{early}. CMT~\cite{cmt} combines Transformer blocks with efficient convolution operators like depthwise convolution~\cite{mobilenetv2}, and achieves better efficiency-performance trade-off. ACmix~\cite{acmix} shares the computation overhead of convolution and self-attention, and integrates both modules with limited cost. Methods have also been proposed for the efficient training of Transformers~\cite{efficient_train,deep_incubation}. In application scenarios demanding high efficiency, MobileFormer~\cite{mobileformer} maintains two paths for convolution and Transformer respectively and enjoys the benefit from both modules. Dyn-Perceiver~\cite{dyn_perceiver} achieves efficient visual recognition through dynamic early exiting~\cite{l2w_den,han2021dynamic,yang2020resolution}. MobileViT~\cite{1mobilevit} takes advantage of the success of MobileNets~\cite{mobilenetv2} and uses the combination of mobilenet blocks and Transformer blocks to achieve light-weight and low latency.

However, these approaches still relied on the Softmax operator, whose inherit high computation complexity inevitably results in the inconvenience in model architecture design and practical application.

\subsection{Linear Attention}
Apart from the above methods, another line of research addresses high computation complexity with linear attention~\cite{linear_attn}. Specifically, linear attention replaces the Softmax function in self-attention with separate kernel functions. In this case, linear attention does not have to compute the pairwise similarity $QK^T$ first. As illustrated in Fig.~\ref{fig:softmax_linear_attn}, based on the associative property of matrix multiplication, linear attention can change the computation order by computing $K^TV$ first, thus reducing the computation complexity from $\mathcal{O}(N^2d)$ to $\mathcal{O}(Nd^2)$. Though efficient, how to design linear attention module as effective as softmax attention is a nontrivial problem. Performer~\cite{performer} approximates the Softmax operation with
orthogonal random features. Efficient attention~\cite{efficient_attn} applies Softmax function to $Q$ and $K$ respectively, which naturally ensures each row of $QK^T$ sums up to 1. Nyströmformer~\cite{nystromformer} and SOFT~\cite{soft} approximate the full self-attention matrix via matrix decomposition. Hydra attention~\cite{hydra_attn} replaces Softmax with cosine similarity and proposes hydra trick which reduces the computation complexity to $\mathcal{O}(Nd)$. EfficientVit~\cite{efficientvit} uses depth-wise convolution to improve linear attention's local feature extraction capacity. Castling-ViT~\cite{castling_vit} proposes linear angular kernel to measure spectral similarity between each $Q_i$ and $K_j$.

\begin{figure}[t]
    \centering
    \includegraphics[width=0.95\linewidth]{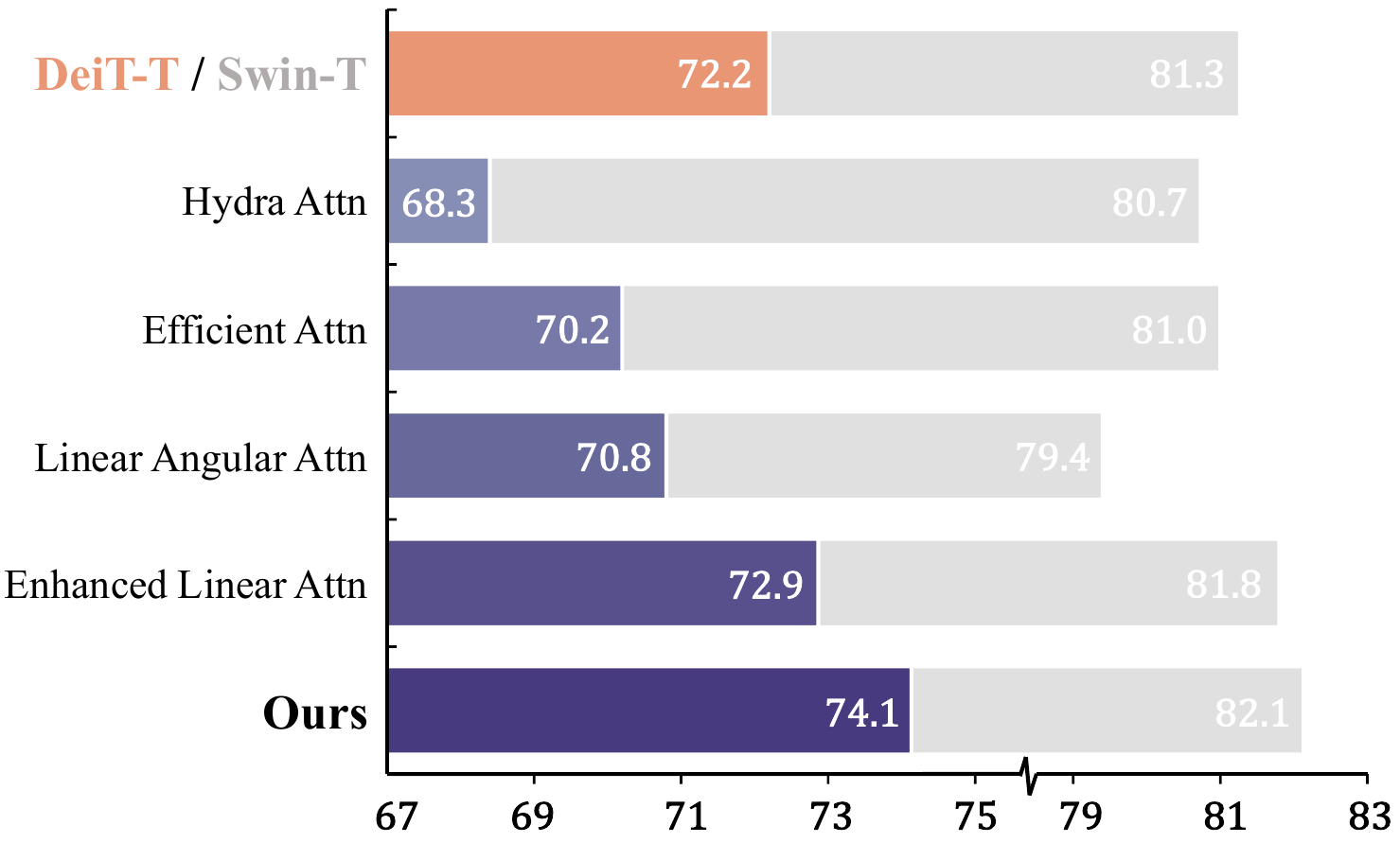}
    \caption{Comparison of different linear attention designs on DeiT-Tiny and Swin-Tiny structures.}
    \label{fig:other_la}
    \vskip -0.05in
\end{figure}

Nevertheless, current linear attention designs either do not have enough expressive capability to catch up with Softmax attention or involve additional computation overhead from the complex kernel function. In this work, we analyze the reasons for the performance drop of linear attention from the focus ability and feature diversity perspectives. Based on these analyses, we propose a novel linear attention module called focused linear attention which achieves better performance than Softmax attention with lower computation complexity (\cref{fig:other_la}).

\section{Preliminaries} \label{sec:method}

\subsection{Vision Transformer and Self-Attention} \label{sec:preliminaries}

We first revisit the general form of self-attention in Vision Transformers. Given the input N tokens $ x\in\mathbb{R}^{N \times C} $, within each head, self-attention can be written as:
\begin{equation} \label{eq:general_attn}
    \begin{split}
        Q=xW_Q, K=xW_K, V=xW_V, \\
        O_i=\sum_{j=1}^{N}\ \frac{{\rm Sim}{\left(Q_i,K_j\right)}}{\sum_{j=1}^{N}\ {\rm Sim}{\left(Q_i,K_j\right)}}V_j,
    \end{split}
\end{equation}
where $ W_Q, W_K, W_V\!\in\!\mathbb{R}^{C \times C} $ are projection matrices and $ {\rm Sim}{\left(\cdot,\cdot \right)} $ denotes the similarity function. Modern vision Transformers mainly adopt Softmax attention~\cite{attention} where similarity is measured as $ {\rm Sim}\left(Q,K\right)\!=\!{\rm exp}({QK^T}/{\sqrt d}) $. In this case, the attention map is obtained by computing the similarity between \textit{all} query-key pairs, which leads to the computation complexity of $\mathcal{O}(N^2)$. 

Due to the quadratic computation complexity, simply using self-attention with global receptive field becomes intractable, which usually leads to excessive computation costs. Previous works either addressed this concern by designing sparse global attention pattern~\cite{pvt,dat} or applying smaller attention windows~\cite{swin,cswin}. Though effective, these approaches become susceptible to the carefully-designed attention patterns, or inevitably sacrifice the ability to model long-range dependencies.

\subsection{Linear Attention}
Comparably, linear attention~\cite{linear_attn} is considered as an effective alternative which restricts the computation complexity from $\mathcal{O}(N^2)$ to $\mathcal{O}(N)$. Specifically, carefully designed kernels are introduced as the approximation of the original similarity function, \textit{i.e.,}
\begin{equation}
    {\rm Sim}\left(Q,K\right)=\phi(Q)\phi(K)^T,
\end{equation}
where the self-attention module can be rewritten as:
\begin{equation} \label{eq:linear_attn}
        O_i=\sum_{j=1}^{N}\ \frac{\phi\left(Q_i\right)\phi\left(K_j\right)^T}{\sum_{j=1}^{N}{\phi\left(Q_i\right)\phi\left(K_j\right)^T\ }}V_j. 
\end{equation}
In this way, we can change the computation order from $(QK^T)V$ to $Q(K^TV)$ based on the associative property of matrix multiplication (as illustrated in \cref{fig:softmax_linear_attn}):
\begin{equation} \label{eq:linear_attn}
        O_i=\frac{\phi(Q_i)\left(\sum_{j=1}^{N}{\phi{(K_j)}^TV_j\ }\right)}{\phi(Q_i)\left(\ \sum_{j=1}^{N}{\phi{(K_j)}^T\ }\right)},
\end{equation}
where the computation complexity with respect to token number is reduced to $ \mathcal{O}(N)$.

However, current linear attention approaches also face the dilemma between model complexity and expressiveness. On one hand, simple approximations, \textit{e.g.,} using ReLU activation~\cite{efficientvit}, are too loose and lead to significant performance drop. On the other hand, carefully designed kernel functions~\cite{performer} or matrix decomposition approaches~\cite{soft,nystromformer} may incur additional computation overhead. In general, there is still a gap between the practical performance of linear attention and Softmax attention.



\section{Focused Linear Attention} \label{sec:focused_linear_attention}
Although enjoying linear computational complexity, various previous works have also proved that simply replacing Softmax attention with linear attention usually results in severe performance drop \cite{cosformer,efficientvit,performer,luna}. 
In this section, we first perform a detailed analysis of the inferior performances of linear attention from two perspectives: focus ability and feature diversity. Then, we introduce our \textbf{Focused Linear Attention} which adequately addresses these concerns and achieves high efficiency and expressive capability.

\subsection{Focus ability}

\begin{figure}
    \centering
    \includegraphics[width=\linewidth]{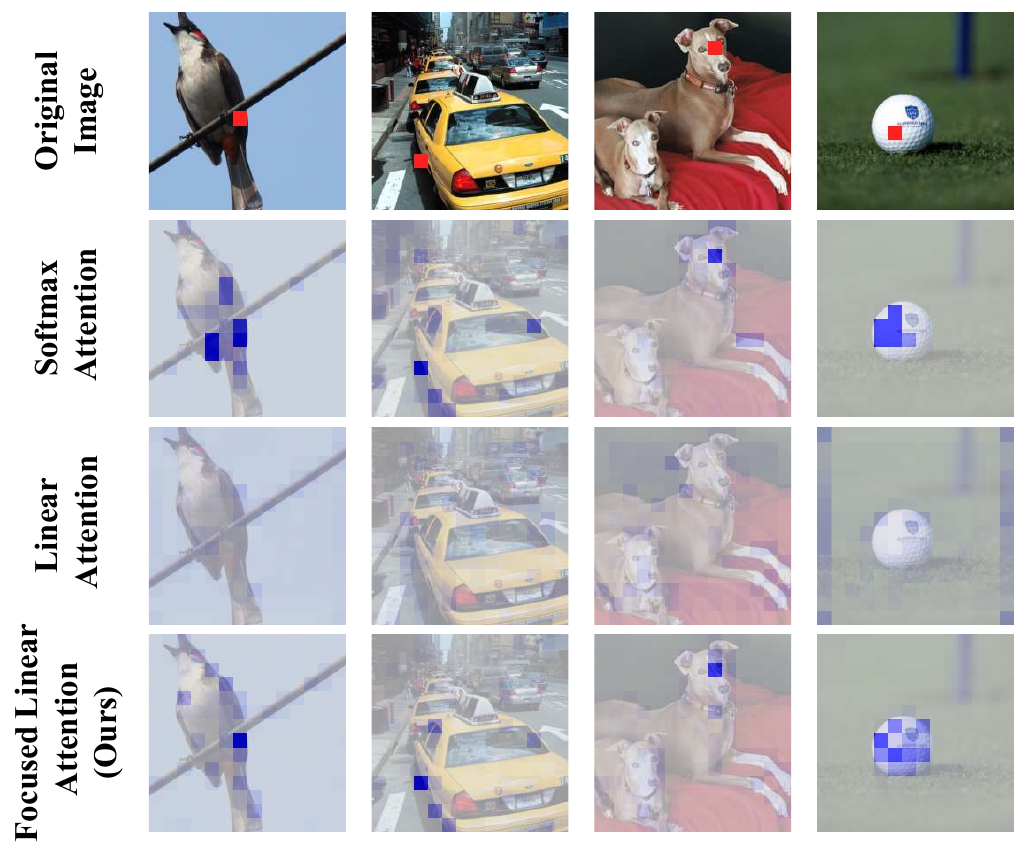}
    \caption{The distribution of Softmax attention, linear attention and our focused linear attention from DeiT-tiny. Softmax attention can produce sharp distribution, while linear attention's distribution is relatively smooth. Our module restores the sharp distribution as the original Softmax attention. Feature corresponding to the red block is used as query. See more visualizations in Appendix. }
    \label{fig:attn_mask}
    \vskip -0.2in
\end{figure}

Softmax attention practically provides a nonlinear re-weighting mechanism, which makes it easy to concentrate on important features~\cite{cosformer,efficientvit,explicit_sparse_transformer}. As shown in \cref{fig:attn_mask}, the distribution of attention map from Softmax attention is especially sharp on certain regions, \textit{e.g.,} foreground objects. Comparably, the distribution in linear attention is relatively smooth, making its output closer to the average of all features and failing to focus on more informative regions.

As a remedy, we propose a simple yet effective solution by adjusting the direction of each query and key features, driving similar query-key pairs closer while pushing dissimilar query-key pairs away. Specifically, we present a simple mapping function $f_p$ called \textbf{Focused Function}:
\begin{align}\label{eq:focus_function}
        &{\rm Sim}\left(Q_i,K_j\right)=\phi_p\left(Q_i\right){\phi_p\left(K_j\right)}^T, \\
        {\rm where} \ \ \phi_p(x)&\!\!=\!\!f_p\left({\rm ReLU}(x)\right), \ f_p(x)\!\!=\!\!\frac{\left\| x\right\|}{\left\| x^{\ast\ast p}\right\|} x^{\ast\ast p},
\end{align}
and $ x^{\ast\ast p} $ represents element-wise power $ p $ of $ x $. We follow previous linear attention modules to use the ReLU function first to ensure the non-negativity of input and validity of denominator in Eq.(\ref{eq:linear_attn}).
A direct observation is that the norm of the feature is preserved after the mapping, \textit{i.e.,} $\left\|x\right\|\! =\! \left\|f_p(x)\right\|$, indicating that only feature direction is adjusted.

On this basis, we show that under mild assumptions, the proposed mapping function $f_p$ practically affects the distribution of attention.

\noindent
\textbf{Proposition 1} (Feature direction adjustment with $f_p$) \textit{Let}
$x\!=\!\left(x_1,\cdots,x_n\right), y\!=\!\left(y_1,\cdots,y_n\right)\!\in\!\mathbb{R}^n, x_i, y_j \!\ge\! 0$. \textit{Assume $x$ and $y$ have the \textbf{single} largest value $x_m$ and $y_n$ respectively. For a pair of feature $\{x,y\}$ with $m\!=\!n$:}
\begin{equation}\label{eq:similar_bigger_method}
    \exists \ p>1, \ s.t.\ \left\langle \phi_{p}(x), \phi_{p}(y)\right\rangle > \left\langle x, y\right\rangle. 
\end{equation}
\textit{For a pair of feature $\{x,y\}$ with $m\!\neq\!n$:}
\begin{equation}\label{eq:dissimilar_smaller_method}
    \exists \ p>1, \ s.t.\ \left\langle \phi_{p}(x), \phi_{p}(y)\right\rangle < \left\langle x, y\right\rangle. 
\end{equation}

\begin{proof} 
Please refer to Appendix for complete proof.
\end{proof}
Therefore, with a proper $p$, 
our focused function $f_p(\cdot)$ practically achieves a more distinguished difference between similar query-key pairs (\cref{eq:similar_bigger_method}) and dissimilar query-key pairs (\cref{eq:dissimilar_smaller_method}), restoring the sharp attention distribution as the original Softmax function.

\begin{figure}
    \centering
    \includegraphics[width=\linewidth]{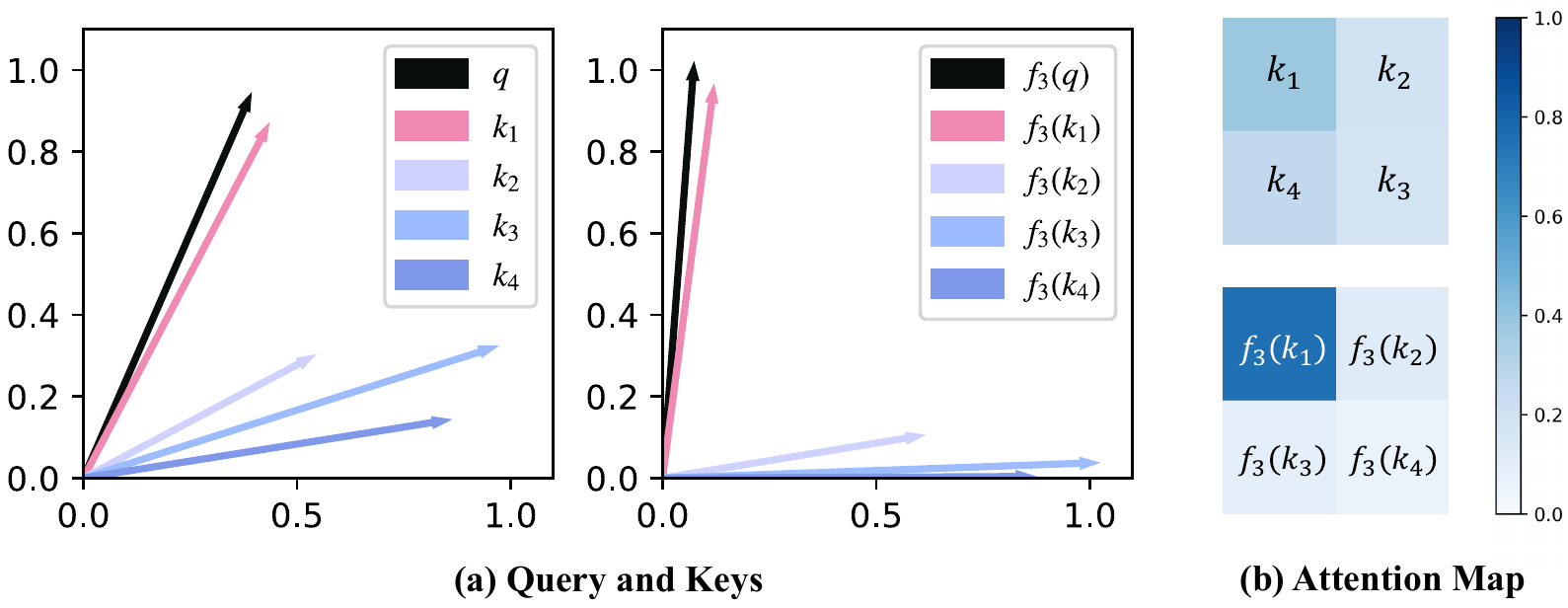}
    \caption{(a) $ f_p $ ``pulls" each vector to its nearest axis, thus helping linear attention focus on similar features. (b) The vanilla linear attention scores are $ [0.37, 0.19, 0.26, 0.18] $, while the attention scores after $ f_3 $ are $ [0.75, 0.11, 0.09, 0.05]. $ }
    \label{fig:fp_effect}
\end{figure}

For better understanding, we give an example to show the effects of $ f_p $ in \cref{fig:fp_effect}. It can be seen that $ f_p $ actually ``pulls" each vector to its nearest axis, and $ p $ determines the degree of this ``pulling". By doing so, $ f_p $ helps divide the features into several groups according to their nearest axes, improving the similarity within each group while reducing the similarity between the groups. The visualizations are in accordance with our analysis above.


\subsection{Feature diversity}

\begin{figure}
    \centering
    \includegraphics[width=\linewidth]{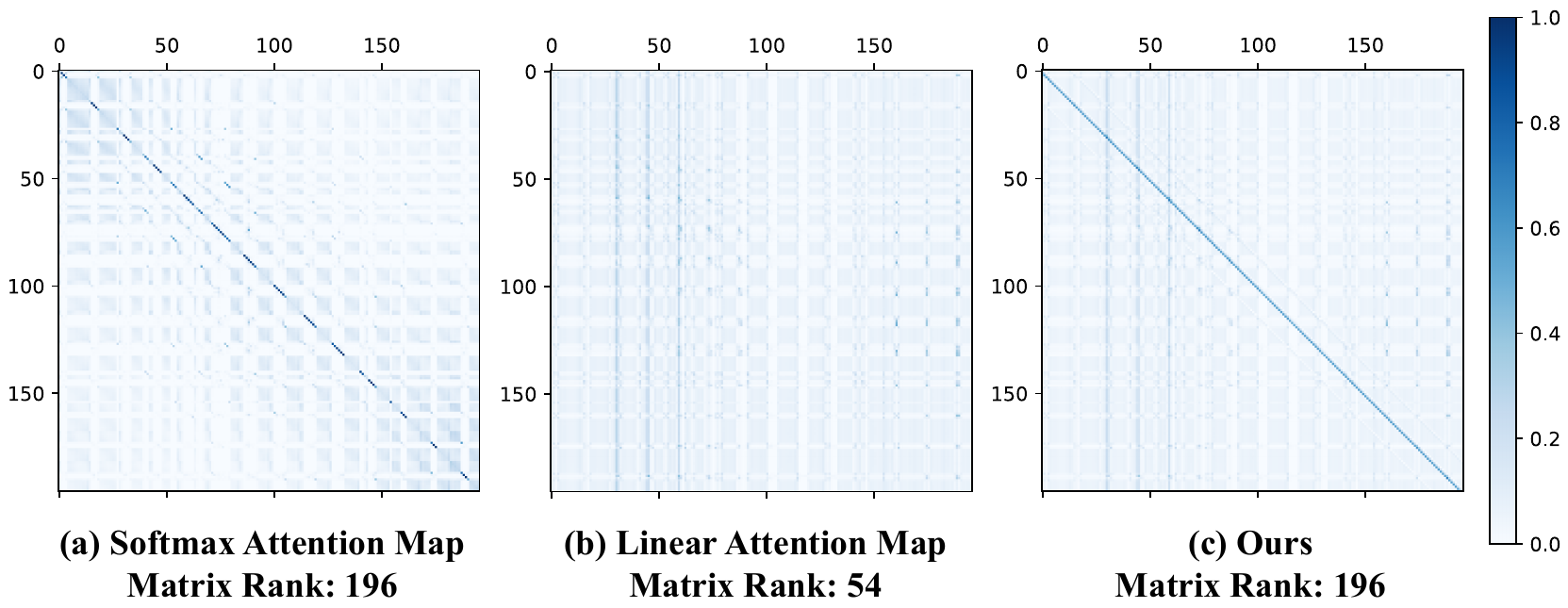}
    \caption{Attention map ($ 196\!\times\! 196 $) from the 3rd block of DeiT-tiny. (a) Softmax attention can learn a full-rank attention map. (b) Linear attention can not learn an attention map with a rank greater than head dim 64. Many rows of the attention map are seriously homogenized, resulting in the resemblance among output features. (c) The lightweight DWC helps linear attention learn an equivalent attention map with a high rank and maintain feature diversity. Both (b) and (c) involve focused function $f_p$. }
    \label{fig:attn_rank}
\end{figure}

Apart from focus ability, feature diversity is also one of the factors that set restriction on the expressive power of linear attention. One of the possible reasons may give credit to the rank of the attention matrix~\cite{combiner,paramixer}, where a significant difference can be seen. Take one of the Transformer layers from DeiT-Tiny~\cite{deit} with $N\!=\!14\!\times\!14$ for example, we can see from \cref{fig:attn_rank} (a) that the attention matrix has the full rank (196 out of 196), showing the diversity when aggregating features from values.

Nevertheless, this can be hardly achieved in the case of linear attention. As a matter of fact, the rank of the attention matrix in linear attention is bounded by the number of tokens $N$ and the channel dimension $d$ for each head:
\begin{align} \label{eq:rank_problem}
    {\rm rank}(\phi(Q)\phi(K)^T)&\!\le\! {\rm min}\{{\rm rank}(\phi(Q)),{\rm rank}(\phi(K))\} \nonumber \\
    &\!\le\! {\rm min}\{N,d\},
\end{align}
where $d$ is usually smaller than $N$ in common vision Transformer designs, \textit{e.g.,} $d\!=\!64,N\!=\!196$ in DeiT~\cite{deit} and $d\!=\!32,N\!=\!49$ in Swin Transformer~\cite{swin}. In this case, the upper bound of attention matrix rank is restricted at a lower ratio, which indicates that many rows of the attention map are seriously homogenized. As the output of self-attention is the weighted sum of the same set of $V$, the homogenization of attention weights inevitably leads to the resemblance among the aggregated features.

To better illustrate, we substitute the original Softmax attention in DeiT-Tiny with linear attention, and show the rank of the attention map in \cref{fig:attn_rank} (b). It can be observed that the rank is greatly decreased (54 out of 196) and many rows of the attention matrix are similar.

As a remedy, we present a simple yet effective solution to address this limitation of linear attention. Specifically, a depthwise convolution (DWC) module is added to the attention matrix and the output can be formulated as:
\begin{equation}\label{eq:linear_attn_dwc}
    O=\phi(Q)\phi(K)^TV+{\rm DWC}\left(V\right).
\end{equation}
To better understand the effect of this DWC module, we can consider it as a kind of attention, in which each query will only focus on several adjacent features in space instead of all features $V$. This locality ensures that even if the linear attention values corresponding to two queries are the same, we can still get different outputs from different local features, thus maintaining feature diversity. The effect of DWC can also be explained from the perspective of matrix rank. Based on Eq.(\ref{eq:linear_attn_dwc}), we have:
\begin{equation}
    O=\left(\phi(Q)\phi(K)^T+M_{\rm DWC}\right)V=M_{eq}V,
\end{equation}
where we denote $ M_{\rm DWC}$ as the sparse matrix corresponding to the depthwise convolution function, and denote $ M_{eq} $ as the equivalent full attention map. As $ M_{\rm DWC}$ has the potential to be a full rank matrix, we practically 
increase the upper bound of the rank of the equivalent attention matrix, which incurs little computation overhead while greatly improving the linear attention’s performance.

To better illustrate, we conduct similar modifications on DeiT-Tiny. With the additional DWC module, the rank of the attention map in the linear attention can be restored to full rank (196 out 196 as shown in \cref{fig:attn_rank} (c)), which keeps the feature diversity as the original Softmax attention.



\begin{figure*}[t]
    \begin{minipage}{0.6\linewidth}
        \centerline{\includegraphics[width=0.95\linewidth]{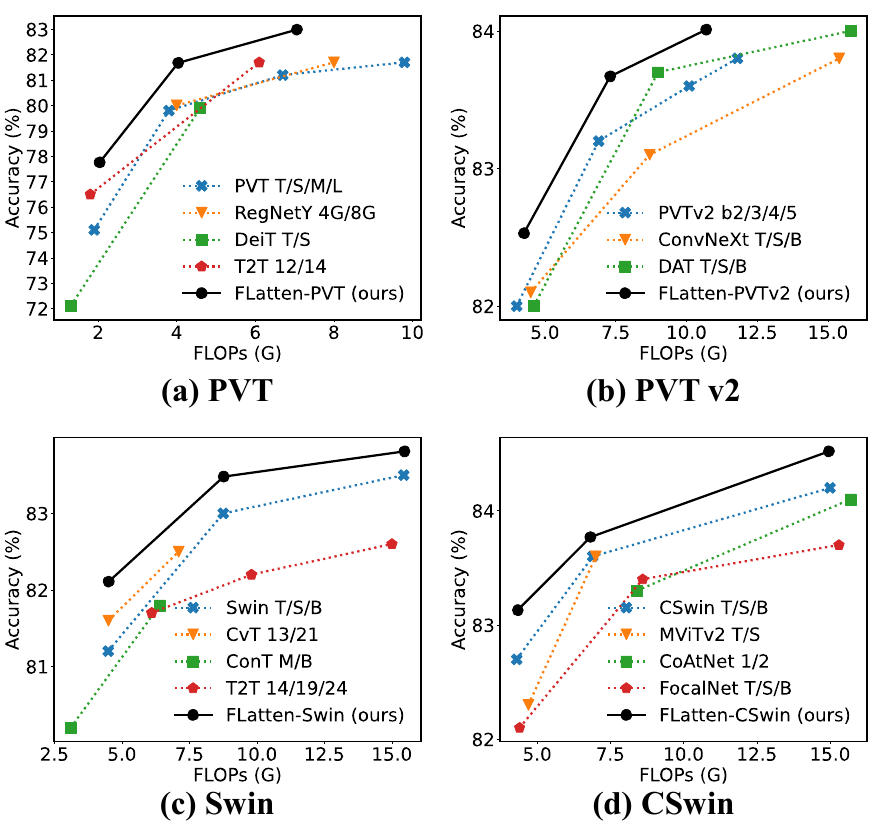}}
    \end{minipage}
    \hfill
    \begin{minipage}{0.4\linewidth}\footnotesize
        \renewcommand\arraystretch{0.85}
        \setlength{\tabcolsep}{0.95mm}{
        \begin{tabular}{l|c c c|l}
            \toprule
            \textbf{Method} 
            & \textbf{Reso}   & \textbf{\#Params} & \textbf{Flops}    & \textbf{Top-1}\\
            
            \midrule
            DeiT-T~\cite{deit}  
            & ${224}^2$     & 5.7M     & 1.2G      & 72.2\\
            \rowcolor{lightgray!50} \textbf{FLatten-DeiT-T} 
            & ${224}^2$     & 6.1M     & 1.1G      & \textbf{74.1\,{\scriptsize (+1.9)}}\\
            
            \midrule
            PVT-T~\cite{pvt}  
            & ${224}^2$     & 13.2M     & 1.9G      & 75.1\\
            \rowcolor{lightgray!50} \textbf{FLatten-PVT-T} 
            & ${224}^2$     & 12.2M     & 2.0G      & \textbf{77.8\,{\scriptsize (+2.7)}}\\
            PVT-S 
            & ${224}^2$     & 24.5M     & 3.8G      & 79.8\\
            \rowcolor{lightgray!50} \textbf{FLatten-PVT-S} 
            & ${224}^2$     & 21.7M     & 4.0G      & \textbf{81.7\,{\scriptsize (+1.9)}}\\
            
            \midrule
            PVTv2-B1~\cite{pvtv2}  
            & ${224}^2$     & 13.1M     & 2.1G      & 78.7\\
            \rowcolor{lightgray!50} \textbf{FLatten-PVTv2-B1} 
            & ${224}^2$     & 12.9M     & 2.2G      & \textbf{79.5\,{\scriptsize (+0.7)}}\\
            PVTv2-B2  
            & ${224}^2$     & 25.4M     & 4.0G      & 82.0\\
            \rowcolor{lightgray!50} \textbf{FLatten-PVTv2-B2} 
            & ${224}^2$     & 22.6M     & 4.3G      & \textbf{82.5\,{\scriptsize (+0.5)}}\\
            
            \midrule
            Swin-T~\cite{swin}  
            & ${224}^2$     & 29M       & 4.5G      & 81.3\\
            \rowcolor{lightgray!50} \textbf{FLatten-Swin-T} 
            & ${224}^2$     & 29M       & 4.5G      & \textbf{82.1\,{\scriptsize (+0.8)}}\\
            Swin-S 
            & ${224}^2$     & 50M       & 8.7G      & 83.0\\
            \rowcolor{lightgray!50} \textbf{FLatten-Swin-S} 
            & ${224}^2$     & 51M       & 8.7G      & \textbf{83.5\,{\scriptsize (+0.5)}}\\
            Swin-B 
            & ${224}^2$     & 88M       & 15.4G     & 83.5\\
            \rowcolor{lightgray!50} \textbf{FLatten-Swin-B} 
            & ${224}^2$     & 89M       & 15.4G     & \textbf{83.8\,{\scriptsize (+0.3)}}\\
            Swin-B 
            & ${384}^2$     & 88M       & 47.0G     & 84.5\\
            \rowcolor{lightgray!50} \textbf{FLatten-Swin-B} 
            & ${384}^2$     & 91M       & 46.5G     & \textbf{85.0\,{\scriptsize (+0.5)}}\\
            
            \midrule
            CSwin-T~\cite{cswin} 
            & ${224}^2$     & 23M       & 4.3G      & 82.7\\
            \rowcolor{lightgray!50} \textbf{FLatten-CSwin-T} 
            & ${224}^2$     & 21M       & 4.3G      & \textbf{83.1\,{\scriptsize (+0.4)}}\\
            CSwin-S 
            & ${224}^2$     & 35M       & 6.9G      & 83.6\\
            \rowcolor{lightgray!50} \textbf{FLatten-CSwin-S} 
            & ${224}^2$     & 35M       & 6.9G      & \textbf{83.8\,{\scriptsize (+0.2)}}\\
            CSwin-B 
            & ${224}^2$     & 78M       & 15.0G     & 84.2\\
            \rowcolor{lightgray!50} \textbf{FLatten-CSwin-B} 
            & ${224}^2$     & 75M       & 15.0G     & \textbf{84.5\,{\scriptsize (+0.3)}}\\
            CSwin-B 
            & ${384}^2$     & 78M       & 47.0G     & 85.4\\
            \rowcolor{lightgray!50} \textbf{FLatten-CSwin-B} 
            & ${384}^2$     & 78M       & 46.4G     & \textbf{85.5\,{\scriptsize (+0.1)}}\\
            \bottomrule
        \end{tabular}}
    \end{minipage}
    \vskip 0.1in
    \caption{Comparison of different models on ImageNet-1K. See the full comparison table in Appendix.}
    \label{fig:main}
    \vskip -0.1in
\end{figure*}

\subsection{Focused linear attention module}
Based on the aforementioned analysis, we propose a novel linear attention module, dubbed \emph{focused linear attention}, which reduces the computation complexity while maintaining the expressive power. Specifically, we first design a novel mapping function to imitate the sharp distribution of the original Softmax attention. On this basis, we focus on the low-rank dilemma in previous linear attention modules, and adopt a simple depthwise convolution to restore feature diversity. In this way, our new module can enjoy benefits from both linear complexity and high expressiveness. Specifically, our module can be formulated as:
\begin{equation}
    O\!=\!{\rm Sim}(Q,K)V \!=\!\phi_p(Q){\phi_p(K)}^TV\!+\!{\rm DWC}(V).
\end{equation}

In general, our module has the following advantages:

(1) \textbf{Low computation complexity as linear attention.} By changing the computation order of self-attention, the complexity is transformed from $\mathcal{O}(N^2d)$ to $\mathcal{O}(Nd^2)$, where $N$ and $d$ denote the token number and channel dimension of each head respectively. $d$ is usually smaller than $N$ in common vision Transformer designs, \textit{e.g.,} $d\!=\!64,N\!=\!196$ in DeiT~\cite{deit} and $d\!=\!32,N\!=\!49$ in Swin Transformer~\cite{swin}, the overall computation is practically decreased. Also, compared to previous linear attention modules~\cite{performer} that design complex kernel function, our proposed focused function $ f_p $ only adopts simple operators which achieves approximation with minimum computation overhead.


(2) \textbf{High expressive capability as Softmax attention.} As we have analyzed above, previous kernel-based linear attention designs are generally inferior to the Softmax counterpart from the focus ability and feature diversity perspective. With the proposed focused function $ f_p $ and depthwise convolution, our \emph{focused linear attention} can achieve even better performance than Softmax attention.

In addition, our module also has the potential of adapting to larger receptive field and different model architectures. 
Modern Transformer models based on Softmax attention mainly use a limited number of key/value pairs because of the quadratic complexity towards token numbers. 
Nevertheless, the linear complexity of our module endows us to expand the receptive field to a larger region while maintaining the same amount of computation, and enjoying the advantage of modeling long-range dependencies. Also, our module can serve as a plug-in module and be easily adopted on a variety of modern vision Transformer architectures.
We empirically implement our module on five advanced models including DeiT \cite{deit}, PVT \cite{pvt}, PVT-v2 \cite{pvtv2}, Swin Transformer \cite{swin} and CSwin Transformer \cite{cswin}. Considering the advantage of enlarged receptive field, we adopt the focused linear attention block at early stages of the vision Transformers, and keep the rest of blocks unchanged. Detailed model architectures are shown in Appendix.

\section{Experiments} \label{sec:experiments}
To verify the effectiveness of our method, we conduct experiments on ImageNet-1K classification~\cite{imagenet}, ADE20K semantic segmentation~\cite{ade20k}, and COCO object detection~\cite{coco}. We also provide a detailed comparison with other linear attention modules based on two representative model structures. In addition, we perform comprehensive ablation studies to analyze each important design element.

\subsection{ImageNet-1K Classification}
ImageNet-1K~\cite{imagenet} contains 1.28M images for training and 50K images for validation. We practically implement our module on five advanced Vision Transformer models, and report the Top-1 accuracy on the validation split to compare with various state-of-the-art models.

\begin{table}[t]
\vskip 0.2in
\newcommand{\tabincell}[2]{\begin{tabular}{@{}#1@{}}#2\end{tabular}}
\begin{center}
\setlength{\tabcolsep}{0.65mm}{
\renewcommand\arraystretch{1.1}
\begin{tabular}{l|c|cc|cc}
\toprule
\multicolumn{6}{c}{\textbf{Semantic Segmentation on ADE20K}} \\
Backbone & Method & FLOPs & \#Params & mIoU & mAcc \\
\hline
PVT-T & S-FPN & 158G & 17M & 36.57 & 46.72 \\
\rowcolor{lightgray!50}
\textbf{FLatten-PVT-T} & S-FPN & 169G & 16M & \textbf{37.21} & 48.95 \\
\hline
Swin-T & UperNet & 945G & 60M & 44.51 & 55.61 \\
\rowcolor{lightgray!50}
\textbf{FLatten-Swin-T} & UperNet & 946G & 60M & \textbf{44.82} & 57.01 \\
\hline
Swin-S & UperNet & 1038G & 81M & 47.64 & 58.78 \\
\rowcolor{lightgray!50}
\textbf{FLatten-Swin-S} & UperNet & 1038G & 82M & \textbf{48.14} & 59.31 \\
\toprule
\end{tabular}}
\end{center}
\vskip -0.1in
\caption{Results of semantic segmentation. The FLOPs are computed over encoders and decoders with an input image at the resolution of 512$\times$2048. S-FPN is short for SemanticFPN \cite{semfpn} model.}
\label{tab:seg}
\vskip -0.2in
\end{table}

\begin{table*}[t]
\begin{center}
\setlength{\tabcolsep}{1.0mm}{
\renewcommand\arraystretch{1.1}
\begin{tabular}{l|c|c|c|ccc|ccc|ccc|ccc}
    \toprule
    \multicolumn{16}{c}{\textbf{(a) Mask R-CNN Object Detection \& Instance Segmentation on COCO}} \\
    Method & FLOPs & \#Param & Schedule & AP$^b$ & AP$^b_\text{50}$ & AP$^b_\text{75}$ & AP$^b_s$ & AP$^b_m$ & AP$^b_l$ & AP$^m$ & AP$^m_\text{50}$ & AP$^m_\text{75}$ & AP$^m_s$ & AP$^m_m$ & AP$^m_l$ \\
    
    \hline PVT-T 
    & 240G & 33M  & 1x      & 36.7 & 59.2 & 39.3 
    & 21.6 & 39.2 & 49.0    & 35.1 & 56.7 & 37.3    & 19.5 & 37.4 & 48.5 \\
    \rowcolor{lightgray!50} \textbf{FLatten-PVT-T} 
    & 244G & 32M  & 1x      & 38.2 & 61.6 & 41.9    
    & 24.1 & 40.7 & 51.0    & 37.0 & 57.6 & 39.0    & 19.4 & 39.0 & 52.1 \\
    
    
    \hline Swin-T 
    & 267G & 48M  & 1x      & 43.7 & 66.6 & 47.7 
    & 28.5 & 47.0 & 57.3    & 39.8 & 63.3 & 42.7    & 24.2 & 43.1 & 54.6 \\
    \rowcolor{lightgray!50} \textbf{FLatten-Swin-T} 
    & 268G & 49M  & 1x      & 44.2 & 67.3 & 48.5 
    & 29.4 & 47.5 & 57.0    & 40.2 & 63.8 & 43.0    & 24.5 & 43.8 & 54.7 \\
    
    \hline Swin-T 
    & 267G & 48M  & 3x      & 46.0 & 68.1 & 50.3 
    & 31.2 & 49.2 & 60.1    & 41.6 & 65.1 & 44.9    & 25.9 & 45.1 & 56.9 \\
    \rowcolor{lightgray!50} \textbf{FLatten-Swin-T} 
    & 268G & 49M  & 3x      & 46.5 & 68.5 & 50.8 
    & 31.2 & 49.6 & 60.4    & 42.1 & 65.4 & 45.1    & 25.4 & 45.4 & 56.8 \\
    \toprule
    
    \multicolumn{16}{c}{\textbf{(b) Cascade Mask R-CNN Object Detection \& Instance Segmentation on COCO}} \\
    Method & FLOPs & \#Param & Schedule & AP$^b$ & AP$^b_\text{50}$ & AP$^b_\text{75}$ & AP$^b_s$ & AP$^b_m$ & AP$^b_l$ & AP$^m$ & AP$^m_\text{50}$ & AP$^m_\text{75}$ & AP$^m_s$ & AP$^m_m$ & AP$^m_l$ \\
    
    \hline Swin-T 
    & 745G & 86M  & 3x      & 50.4 & 69.2 & 54.7 
    & 33.8 & 54.1 & 65.2    & 43.7 & 66.6 & 47.3    & 27.3 & 47.5 & 59.0 \\
    \rowcolor{lightgray!50} \textbf{FLatten-Swin-T} 
    & 747G & 87M  & 3x      & 50.8 & 69.6 & 55.1 
    & 34.2 & 54.6 & 65.5    & 44.1 & 67.0 & 48.1    & 27.6 & 48.1 & 59.0 \\
    
    \hline Swin-S 
    & 838G & 107M & 3x      & 51.9 & 70.7 & 56.3 
    & 35.2 & 55.7 & 67.7    & 45.0 & 68.2 & 48.8    & 28.8 & 48.7 & 60.6 \\
    \rowcolor{lightgray!50} \textbf{FLatten-Swin-S} 
    & 841G & 108M & 3x      & 52.2 & 71.2 & 56.8 
    & 35.6 & 56.4 & 67.6    & 45.4 & 68.3 & 49.4    & 29.3 & 49.0 & 60.8 \\
    
    \toprule
\end{tabular}}
\end{center}
\vskip -0.1in
\caption{Results on COCO dataset. The FLOPs are computed over backbone, FPN and detection head with input resolution of 1280$\times$800. }
\label{tab:det2}
\vskip -0.05in
\end{table*}

For fair comparison, we use the exact same settings as the corresponding baseline model to train our FLatten model. Specifically, we use AdamW~\cite{adamw} optimizer to train all our models for 300 epochs with a cosine learning rate decay and 20 epochs of linear warm-up. The basic learning rate for a batch size of 1024 is set to $1\times{10}^{-3}$, and then linearly scaled \textit{w.r.t.} the batch size. We follow DeiT~\cite{deit} and apply RandAugment~\cite{randaugment}, Mixup~\cite{mixup}, CutMix~\cite{cutmix} and random erasing~\cite{random_erasing} to avoid overfitting. In addition, a weight decay of 0.05 is used. To be consistent with \cite{cswin}, we also adopt EMA~\cite{ema} in the training of our FLatten-CSwin models. In terms of larger resolution finetuning, we follow the setting in~\cite{swin, cswin} that finetunes the models for 30 epochs.

The classification results are provided in \cref{fig:main}. It is shown that our method achieves consistent improvements against baseline models under comparable FLOPs or parameters. For example, our FLatten-PVT-T/S surpass PVT-T/S by 2.7\% and 1.9\% respectively with similar FLOPs. Based on Swin, our model achieves comparable performance with 60\% FLOPs. Our model based on PVT-v2 and CSwin also achieves a better trade-off between computation cost and model performance. These results demonstrate that our module has high expressive capability and is applicable to various model structures.

\begin{table}[]
    \centering
        \setlength{\tabcolsep}{1.8mm}{
        \renewcommand\arraystretch{1.2}
        \begin{tabular}{c|c c|c}
            \bottomrule
            \multicolumn{4}{c}{\textbf{(a) Comparison on DeiT-T Setting}} \\
            Linear Attention            & FLOPs     & \#Param   & Acc. \\
            \hline
            Hydra Attn \cite{hydra_attn}
                                        & 1.1G      & 5.7M      & 68.3 \\
            Efficient Attn \cite{efficient_attn}
                                        & 1.1G      & 5.7M      & 70.2 \\
            Linear Angular Attn \cite{castling_vit}
                                        & 1.1G      & 5.7M      & 70.8 \\
            Enhanced Linear Attn \cite{efficientvit}
                                        & 1.1G      & 5.8M      & 72.9 \\
            \hline \rowcolor{lightgray!50}
            \textbf{Ours}               & 1.1G      & 6.1M      & \textbf{74.1} \\
            \toprule

            \multicolumn{4}{c}{\textbf{(b) Comparison on Swin-T Setting}} \\
            Linear Attention            & FLOPs     & \#Param   & Acc. \\
            \hline
            Hydra Attn \cite{hydra_attn}
                                        & 4.5G      & 29M       & 80.7 \\
            Efficient Attn \cite{efficient_attn}
                                        & 4.5G      & 29M       & 81.0 \\
            Linear Angular Attn \cite{castling_vit}
                                        & 4.5G      & 29M       & 79.4 \\
            Enhanced Linear Attn \cite{efficientvit}
                                        & 4.5G      & 29M       & 81.8 \\
            \hline
            \rowcolor{lightgray!50}
            \textbf{Ours}               & 4.5G      & 29M       & \textbf{82.1} \\
            \toprule
        \end{tabular}}
    
    \caption{Comparison of different linear attention designs on DeiT-Tiny and Swin-Tiny structures.}
    \label{tab:other_la}
    \vskip -0.1in
\end{table}

\subsection{Semantic Segmentation}

ADE20K \cite{ade20k} is a widely adopted benchmark for semantic segmentation with 20K/2K training/validation images. We employ our model on two representative segmentation models, SemanticFPN \cite{semfpn} and UperNet \cite{upernet}. As shown in \cref{tab:seg}, our model achieves consistently better results under all settings. Specifically, we can see a $0.5\!\sim\!1\%$ mIoU improvement with comparable computation cost and parameters. The improvements in mAcc are more significant.

\subsection{Object Detection}

COCO \cite{coco} object detection and instance segmentation dataset has 118K training and 5K validation images. We use ImageNet pretrained model as the backbone in Mask R-CNN \cite{mrcn} and Cascade Mask R-CNN \cite{cmrcn} frameworks to evaluate the effectiveness. We conduct experiments on 1x and 3x schedules with different detection heads and show results in \cref{tab:det2}. Taking advantage of larger receptive field, our model shows better results under all settings.

\begin{figure*}
    \centering
    \includegraphics[width=0.94\linewidth]{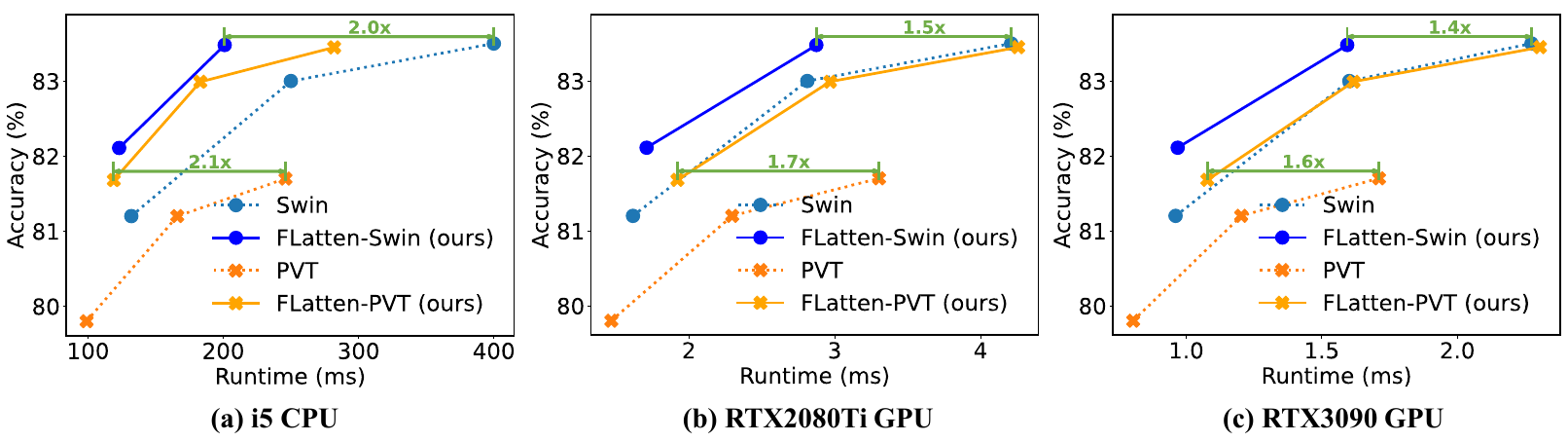}
    \caption{Accuracy-Runtime curve on ImageNet. Runtime is tested with image resolution 224$\!\times\!$224.}
    \label{fig:speed}
    \vskip -0.15in
\end{figure*}

\subsection{Comparison with Other Linear Attention}

To show a fair comparison with other linear attention modules, we conduct experiments based on two representative Vision Transformer structures, DeiT and Swin Transformer respectively. Based on these two models, we compare our focused linear attention module with four previous linear attention designs, including hydra attention~\cite{hydra_attn}, efficient attention~\cite{efficient_attn}, linear angular attention~\cite{castling_vit} and enhanced linear attention~\cite{efficientvit}.

As shown in \cref{tab:other_la}, previous linear attention modules are generally inferior to the Softmax counterpart, while our model significantly outperforms all other designs and the Softmax baseline. This indicates that our module has high expressive capability and can achieve better performance than Softmax attention with lower computation complexity.

\subsection{Inference Time}
We further evaluate the practical efficiency of our model and compare it with two competitive baselines. The results are presented in \cref{fig:speed}. We test the inference latency on multiple hardware platforms, including a desktop CPU (Intel i5-8265U) and two server GPUs (RTX2080Ti and RTX3090). It can be observed that our model achieves a significantly better trade-off between runtime and accuracy on both CPU and GPU, enjoying up to 2.1x faster inference speed with on par or even better performances. 

\subsection{Ablation Study}

\begin{table}[]
    \centering
    \footnotesize
    \setlength{\tabcolsep}{1.5mm}{
    \renewcommand\arraystretch{1.2}
    \begin{tabular}{l|c c|c c}
        \bottomrule
        \                           & FLOPs     & \#Param   & Acc.          & Diff. \\
        \hline
        Vanilla Linear Attention    & 1.1G      & 5.7M      & 70.5          & -3.6 \\
        ~~~ $+$ \ Focused Function  & 1.1G      & 5.7M      & 71.8          & -2.3 \\
        \rowcolor{lightgray!50}
        ~~~ $+$ \ DWC               & 1.1G      & 6.1M      & \textbf{74.1} & \textbf{Ours} \\
        \hline
        DeiT-T                      & 1.2G      & 5.7M      & 72.2          & -1.9 \\
        \toprule
    \end{tabular}}
    \caption{Ablation on each module based on DeiT-T.}
    \label{tab:ablation_focus_dwc}
    \vskip -0.05in
\end{table}
\begin{table}[]
    \centering
    \footnotesize
    \setlength{\tabcolsep}{1.7mm}{
    \renewcommand\arraystretch{1.2}
    \begin{tabular}{c|c c c c c}
        \bottomrule
        focused factor $p$  & 2     & 3     & 4     & 8     & 32  \\
        \hline
        Acc.                & 82.03 & 82.11 & 81.94 & 81.99 & 81.88 \\
        \toprule
    \end{tabular}}
    \caption{Ablation on focused factor p based on FLatten-Swin-T.}
    \label{tab:ablation_on_p}
    \vskip -0.15in
\end{table}

In this section, we ablate the key components in our focused linear attention to verify the effectiveness of these designs. We report the results on ImageNet-1K classification based on FLatten-DeiT-T and FLatten-Swin-T.

\noindent \textbf{Focused function $f_p$ and DWC.}
We first evaluate the effectiveness of our proposed focused function $f_p$ and depth-wise convolution. We start from the vanilla linear attention and introduce $f_p$ and DWC in turn. As shown in \cref{tab:ablation_focus_dwc}, adopting the proposed focused function $f_p$ provides +1.3 improvement. Using DWC to maintain feature diversity further leads to an accuracy gain of +2.3, achieving an overall accuracy of 74.1. These results prove that our proposed $f_p$ and DWC can greatly improve the expressive capability of linear attention, thus helping our focused linear attention module achieve better performance than Softmax attention.

\noindent\textbf{Ablation on different $p$.}
In \cref{tab:ablation_on_p}, we study the effect of focused factor $p$ on the model performance. We find that when $p$ changes between 2 and 32, the Top-1 classification accuracy does not change much, implying the robustness of our module to this hyper-parameter. Practically, we choose $p=3$ for all models in the paper without additional tuning.

\noindent \textbf{Receptive field.}
We also study the impact of receptive field based on FLatten-Swin-tiny. As illustrated in \cref{tab:ablation_window_size}, with the increase of window size, the performance of our model improves progressively. This further proves that though window attention is effective, it inevitably sacrifices the long-range dependency of self-attention from the global perspective and is still inferior to global attention. With linear complexity, it is possible for our module to realize a large even global receptive field while maintaining the same amount of computation.
\begin{table}[]
    \centering
    \footnotesize
    \setlength{\tabcolsep}{1.8mm}{
    \renewcommand\arraystretch{1.2}
    \begin{tabular}{c|c c c|c c}
        \bottomrule
        \                               & Window        & FLOPs & \#Param   & Acc.      & Diff. \\
        \hline
        \multirow{4}{*}{FLatten-Swin-T} & $7^2$         & 4.5G  & 29M       & 81.6      & -0.5 \\
                                        & ${14}^2$      & 4.5G  & 29M       & 81.8      & -0.3 \\
                                        & ${28}^2$      & 4.5G  & 29M       & 81.9      & -0.2 \\
        \rowcolor{lightgray!50}
        \cellcolor{white}               & ${56}^2$      & 4.5G  & 29M       & \textbf{82.1} &\textbf{Ours} \\
        \hline
        Swin-T                          & $7^2$         & 4.5G  & 29M       & 81.3      & -0.8 \\
        \toprule
    \end{tabular}}
    \caption{Ablation on window size based on FLatten-Swin-T.}
    \label{tab:ablation_window_size}
    \vskip -0.1in
\end{table}

\begin{table}[]
    \centering
    \footnotesize
    \setlength{\tabcolsep}{1.1mm}{
    \renewcommand\arraystretch{1.2}
    \begin{tabular}{c c c c|c c|c c}
        \bottomrule
        \multicolumn{4}{c|}{Stages w/ FLatten} & \multirow{2}{*}{FLOPS} & \multirow{2}{*}{\#Param} & \multirow{2}{*}{Acc.} & \multirow{2}{*}{Diff.} \\
        Stage1      & Stage2        & Stage3        & Stage4        &       &       &       &    \\
        \hline
        $\checkmark$&               &               &               & 4.5G  & 29M   & 81.7  & -0.4  \\
        \rowcolor{lightgray!50}
        $\checkmark$&$\checkmark$   &               &               & 4.5G  & 29M   & \textbf{82.1} & \textbf{Ours}  \\
        $\checkmark$&$\checkmark$   &$\checkmark$   &               & 4.5G  & 30M   & 82.0  & -0.1  \\
        $\checkmark$&$\checkmark$   &$\checkmark$   &$\checkmark$   & 4.5G  & 30M   & 81.9  & -0.2  \\
        \hline
        \multicolumn{4}{c|}{Swin-T}                                 & 4.5G  & 29M   & 81.3  & -0.8  \\
        \toprule
    \end{tabular}}
    \caption{Ablation on applying focused linear attention on different stages of the Swin-T structure.}
    \label{tab:ablation_LLSS}
    \vskip -0.2in
\end{table}

\noindent \textbf{Focused linear attention at different stages.}
We replace the shift-window attention of Swin-T with our module at different stages. As shown in \cref{tab:ablation_LLSS}, we can see that replacing the first two stages leads to a performance gain of 0.8, while replacing the last two stages slightly decreases the overall accuracy. We attribute this result to the fact that the first two stages of Swin have larger resolutions and are more suitable for our module with large receptive field.

\section{Conclusion} \label{sec:conclusion}


In this paper, we propose a novel \emph{focused linear attention} module. By addressing the limitations of previous linear attention methods from focus ability and feature diversity perspectives, our module achieves an impressive combination of high efficiency and expressive capability. Extensive experiments on image classification, object detection and semantic segmentation demonstrated that our module can be widely applied to a variety of vision Transformers and achieve a better trade-off between computation efficiency and model performance.

\section*{Acknowledgement}
This work is supported in part by National Key R\&D Program of China (2021ZD0140407), the National Natural Science Foundation of China (62022048, 62276150) and THU-Bosch JCML. We appreciate the generous donation of computing resources by High-Flyer AI.

{\small
\bibliographystyle{ieee_fullname}
\bibliography{_main}
}

\clearpage
\section*{Appendix}
\appendix
\section*{A. Proof of Proposition 1}

As mentioned in the main paper, with the aim to restore the sharp distribution in linear attention, we present our \textbf{Focused Function} $f_p$:
\begin{align}\label{eq:focus_function}
        &{\rm Sim}\left(Q_i,K_j\right)=\phi_p\left(Q_i\right){\phi_p\left(K_j\right)}^T, \\
        {\rm where} \ \ \phi_p(x)&\!\!=\!\!f_p\left({\rm ReLU}(x)\right), \ f_p(x)\!\!=\!\!\frac{\left\| x\right\|}{\left\| x^{\ast\ast p}\right\|} x^{\ast\ast p},
\end{align}
and $ x^{\ast\ast p} $ represents the power $ p $ of $ x $ bit by bit. We follow previous linear attention modules to use the ReLU function first to ensure the non-negativity of input. Therefore, when proving the effects of $f_p$, we only consider $x, y \ge 0$.

\noindent
\textbf{Proposition 1} (Feature direction adjustment with $f_p$) \textit{Let}
$x\!=\!\left(x_1,\cdots,x_n\right), y\!=\!\left(y_1,\cdots,y_n\right)\!\in\!\mathbb{R}^n, x_i, y_j \!\ge\! 0$. \textit{Assume $ 0 < \left\langle x, y \right\rangle < \left\|x\right\| \left\|y\right\| $ and $x$, $y$ have the \textbf{single} largest value $x_m$, $y_n$ respectively. }

\noindent
\textit{For a pair of feature $\{x,y\}$ with $m\!=\!n$:}
\begin{equation}\label{eq:similar_bigger}
    \exists \ p>1, \ s.t.\ \left\langle \phi_{p}(x), \phi_{p}(y)\right\rangle > \left\langle x, y\right\rangle. 
\end{equation}
\textit{For a pair of feature $\{x,y\}$ with $m\!\neq\!n$:}
\begin{equation}\label{eq:dissimilar_smaller}
    \exists \ p>1, \ s.t.\ \left\langle \phi_{p}(x), \phi_{p}(y)\right\rangle < \left\langle x, y\right\rangle. 
\end{equation}

\begin{proof} 

\begin{equation}
    \begin{split}
        \phi_p(x)\!&=\!f_p\left({\rm ReLU}(x)\right)\!=\!f_p(x),\\
        \phi_p(y)\!&=\!f_p\left({\rm ReLU}(y)\right)\!=\!f_p(y).
    \end{split}
\end{equation}
\begin{equation}
    \begin{split}
        \left\| f_p(x)\right\| \!\!=\!\!\frac{\left\| x\right\|}{\left\| x^{\ast\ast p}\right\|} \left\| x^{\ast\ast p} \right\|\!\!=\!\!\left\| x\right\|, \left\| f_p(y)\right\|\!\!=\!\! \left\| y\right\|.
    \end{split}
\end{equation}
Therefore, we have:
\begin{equation}\label{eq:phi_xy_to_uv}
    \begin{split}
        \left\langle \phi_{p}(x), \phi_{p}(y)\right\rangle =& \left\langle f_{p}(x), f_{p}(y)\right\rangle \\
        =& \left\| f_p(x)\right\| \left\| f_p(y)\right\| \left\langle u, v\right\rangle \\
        =& \left\| x\right\| \left\| y\right\| \left\langle u, v\right\rangle,
    \end{split}
\end{equation}
where
\begin{equation}
    \begin{split}
        \left\langle u, v\right\rangle =& \left\langle\frac{f_{p}(x)}{\left\|f_{p}(x)\right\|}, \frac{f_{p}(y)}{\left\|f_{p}(y)\right\|}\right\rangle \\
        =& \frac{\sum_{i=1}^{n} x_{i}^{p} y_{i}^{p}}{\sqrt{\left(\sum_{i=1}^{n} x_{i}^{2 p}\right)\left(\sum_{i=1}^{n} y_{i}^{2 p}\right)}} \\
        =& \frac{\sum_{i=1}^{n} a_{i}^{p} b_{i}^{p}}{\sqrt{\left(\sum_{i=1}^{n} a_{i}^{2 p}\right)\left(\sum_{i=1}^{n} b_{i}^{2 p}\right)}},
    \end{split}
\end{equation}
and 
\begin{equation}
    \begin{split}
        a_{i}=x_{i} / \max _{1 \leq i \leq n}\left(x_{i}\right),& b_{i}=y_{i} / \max _{1 \leq i \leq n}\left(y_{i}\right), \\
        a_{i}, b_{i} &\in[0,1].
    \end{split}
\end{equation}
Based on our assumption, we have:
\begin{equation}
    \exists ! m, \  s.t. \  a_m=1, \ \ \exists ! n,\  s.t.\  b_n=1.
\end{equation}
Therefore,
\begin{equation}
    \lim_{p \rightarrow \infty} a_i^p = 
    \begin{cases}
        1,& \!\!\!i=m \\
        0,& \!\!\!i\neq m \\
    \end{cases}, \ \ \ 
    \lim_{p \rightarrow \infty} b_j^p = 
    \begin{cases}
        1,& \!\!\!j=n \\
        0,& \!\!\!j\neq n \\
    \end{cases}.
\end{equation}
Then we consider the following two cases: \\
\noindent
(1) $m\!=\!n$:
\begin{equation}\label{eq:lim_uv_1}
    \begin{split}
        \lim_{p \rightarrow \infty} \left\langle u, v\right\rangle =& \lim_{p \rightarrow \infty} \frac{\sum_{i=1}^{n} a_{i}^{p} b_{i}^{p}}{\sqrt{\left(\sum_{i=1}^{n} a_{i}^{2 p}\right)\left(\sum_{i=1}^{n} b_{i}^{2 p}\right)}} \\
        =& \  \frac{1\times1}{\sqrt{1\times1}} = 1.
    \end{split}
\end{equation}
\cref{eq:phi_xy_to_uv}, \cref{eq:lim_uv_1} $\Rightarrow$
\begin{equation}
    \begin{split}
        \lim_{p \rightarrow \infty} \left\langle \phi_{p}(x), \phi_{p}(y)\right\rangle =& \lim_{p \rightarrow \infty} \left\| x\right\| \left\| y\right\| \left\langle u, v\right\rangle \\
        =& \left\| x\right\| \left\| y\right\| > \left\langle x, y\right\rangle.
    \end{split}
\end{equation}
Thus we have,
\begin{equation}
    \exists p>1, \ s.t.\ \left\langle \phi_{p}(x), \phi_{p}(y)\right\rangle > \left\langle x, y\right\rangle.
\end{equation}
(2) $m\!\neq\!n$:
\begin{equation}\label{eq:lim_uv_0}
    \begin{split}
        \lim_{p \rightarrow \infty} \left\langle u, v\right\rangle =& \lim_{p \rightarrow \infty} \frac{\sum_{i=1}^{n} a_{i}^{p} b_{i}^{p}}{\sqrt{\left(\sum_{i=1}^{n} a_{i}^{2 p}\right)\left(\sum_{i=1}^{n} b_{i}^{2 p}\right)}} \\
        =& \  \frac{1\times0+0\times1}{\sqrt{1\times1}} = 0.
    \end{split}
\end{equation}
\cref{eq:phi_xy_to_uv}, \cref{eq:lim_uv_0} $\Rightarrow$
\begin{equation}
    \begin{split}
        \lim_{p \rightarrow \infty} \left\langle \phi_{p}(x), \phi_{p}(y)\right\rangle =& \lim_{p \rightarrow \infty} \left\| x\right\| \left\| y\right\| \left\langle u, v\right\rangle \\
        =& \  0 < \left\langle x, y\right\rangle.
    \end{split}
\end{equation}
Thus we have,
\begin{equation}
    \exists p>1, \ s.t.\ \left\langle \phi_{p}(x), \phi_{p}(y)\right\rangle < \left\langle x, y\right\rangle.
\end{equation}

\end{proof}

Therefore, with a proper $p$, our focused function $f_p(\cdot)$ practically achieves a more distinguished difference between similar query-key pairs (\cref{eq:similar_bigger}) and dissimilar query-key pairs (\cref{eq:dissimilar_smaller}). Actually, $ f_p $ divides the features into several groups according to their nearest axes, improving the similarity within each group while reducing the similarity between the groups, thus restoring the sharp attention distribution as the original Softmax function.

\section*{B. More Visualizations}

\begin{figure}
    \vskip -0.03in
    \centering
    \includegraphics[width=\linewidth]{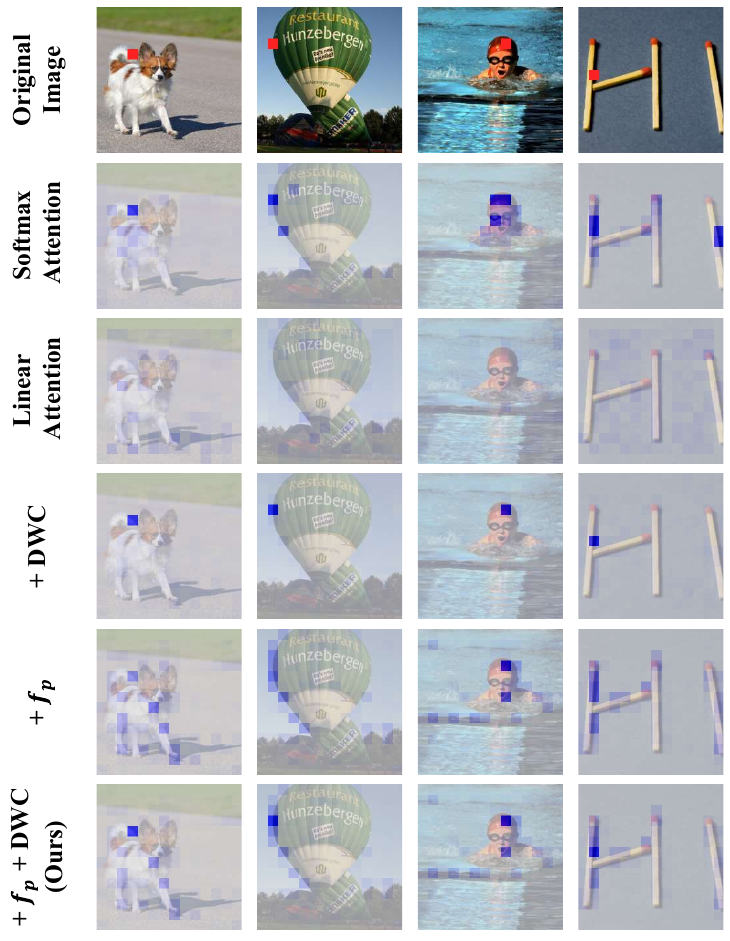}
    \caption{The distribution of attention weights from DeiT-tiny. Feature corresponding to the red block is used as query. }
    \label{fig:attn_mask_more}
    \vskip -0.1in
\end{figure}

We visualize more examples of attention weights in \cref{fig:attn_mask_more}. To better show the contribution of our focused function and DWC, we start from the vanilla linear attention and introduce $f_p$ and DWC separately. As demonstrated in the last three rows, DWC improves local focus ability but cannot focus on any position, while $f_p$ practically enhances model’s focus ability, helping model focus on more informative regions. Combining $f_p$ and DWC, our focused linear attention module restores the sharp distribution as the original Softmax attention.

\section*{C. Full Classification Results}

Due to the page limit, we only present representative ImageNet classification results in Figure 6 of main paper. Here, we give all the classification results when applying our focused linear attention module on various sizes of the five baseline models in Tab.\ref{tab:main_results}.

\begin{table}[]
    \centering
    \renewcommand\arraystretch{1.31}
    \setlength{\tabcolsep}{1.0mm}{
        \begin{tabular}{l|c c c|l}
            \toprule
            \textbf{Method} 
            & \textbf{Reso}   & \textbf{\#Params} & \textbf{Flops}    & \textbf{Top-1}\\
            
            \midrule
            DeiT-T~\cite{deit}  
            & ${224}^2$     & 5.7M     & 1.2G      & 72.2\\
            \rowcolor{lightgray!50} \textbf{FLatten-DeiT-T} 
            & ${224}^2$     & 6.1M     & 1.1G      & \textbf{74.1\,{\scriptsize (+1.9)}}\\
            
            \midrule
            PVT-T~\cite{pvt}  
            & ${224}^2$     & 13.2M     & 1.9G      & 75.1\\
            \rowcolor{lightgray!50} \textbf{FLatten-PVT-T} 
            & ${224}^2$     & 12.2M     & 2.0G      & \textbf{77.8\,{\scriptsize (+2.7)}}\\
            PVT-S 
            & ${224}^2$     & 24.5M     & 3.8G      & 79.8\\
            \rowcolor{lightgray!50} \textbf{FLatten-PVT-S} 
            & ${224}^2$     & 21.7M     & 4.0G      & \textbf{81.7\,{\scriptsize (+1.9)}}\\
            PVT-M
            & ${224}^2$     & 44.2M     & 6.7G      & 81.2\\
            \rowcolor{lightgray!50} \textbf{FLatten-PVT-M} 
            & ${224}^2$     & 37.2M     & 7.0G      & \textbf{83.0\,{\scriptsize (+1.8)}}\\
            PVT-L
            & ${224}^2$     & 61.4M     & 9.8G      & 81.7\\
            \rowcolor{lightgray!50} \textbf{FLatten-PVT-L} 
            & ${224}^2$     & 50.6M     & 10.4G      & \textbf{83.4\,{\scriptsize (+1.7)}}\\
            
            \midrule
            PVTv2-B0~\cite{pvtv2}  
            & ${224}^2$     & 3.4M     & 0.6G      & 70.5\\
            \rowcolor{lightgray!50} \textbf{FLatten-PVTv2-B0} 
            & ${224}^2$     & 3.6M     & 0.6G      & \textbf{71.1\,{\scriptsize (+0.6)}}\\
            PVTv2-B1
            & ${224}^2$     & 13.1M     & 2.1G      & 78.7\\
            \rowcolor{lightgray!50} \textbf{FLatten-PVTv2-B1} 
            & ${224}^2$     & 12.9M     & 2.2G      & \textbf{79.5\,{\scriptsize (+0.7)}}\\
            PVTv2-B2  
            & ${224}^2$     & 25.4M     & 4.0G      & 82.0\\
            \rowcolor{lightgray!50} \textbf{FLatten-PVTv2-B2} 
            & ${224}^2$     & 22.6M     & 4.3G      & \textbf{82.5\,{\scriptsize (+0.5)}}\\
            PVTv2-B3
            & ${224}^2$     & 45.2M     & 6.9G      & 83.2\\
            \rowcolor{lightgray!50} \textbf{FLatten-PVTv2-B3} 
            & ${224}^2$     & 38.3M     & 7.3G      & \textbf{83.7\,{\scriptsize (+0.5)}}\\
            PVTv2-B4
            & ${224}^2$     & 62.6M     & 10.1G      & 83.6\\
            \rowcolor{lightgray!50} \textbf{FLatten-PVTv2-B4} 
            & ${224}^2$     & 51.8M     & 10.7G      & \textbf{84.0\,{\scriptsize (+0.4)}}\\
            
            \midrule
            Swin-T~\cite{swin}  
            & ${224}^2$     & 29M       & 4.5G      & 81.3\\
            \rowcolor{lightgray!50} \textbf{FLatten-Swin-T} 
            & ${224}^2$     & 29M       & 4.5G      & \textbf{82.1\,{\scriptsize (+0.8)}}\\
            Swin-S 
            & ${224}^2$     & 50M       & 8.7G      & 83.0\\
            \rowcolor{lightgray!50} \textbf{FLatten-Swin-S} 
            & ${224}^2$     & 51M       & 8.7G      & \textbf{83.5\,{\scriptsize (+0.5)}}\\
            Swin-B 
            & ${224}^2$     & 88M       & 15.4G     & 83.5\\
            \rowcolor{lightgray!50} \textbf{FLatten-Swin-B} 
            & ${224}^2$     & 89M       & 15.4G     & \textbf{83.8\,{\scriptsize (+0.3)}}\\
            Swin-B 
            & ${384}^2$     & 88M       & 47.0G     & 84.5\\
            \rowcolor{lightgray!50} \textbf{FLatten-Swin-B} 
            & ${384}^2$     & 91M       & 46.5G     & \textbf{85.0\,{\scriptsize (+0.5)}}\\
            
            \midrule
            CSwin-T~\cite{cswin} 
            & ${224}^2$     & 23M       & 4.3G      & 82.7\\
            \rowcolor{lightgray!50} \textbf{FLatten-CSwin-T} 
            & ${224}^2$     & 21M       & 4.3G      & \textbf{83.1\,{\scriptsize (+0.4)}}\\
            CSwin-S 
            & ${224}^2$     & 35M       & 6.9G      & 83.6\\
            \rowcolor{lightgray!50} \textbf{FLatten-CSwin-S} 
            & ${224}^2$     & 35M       & 6.9G      & \textbf{83.8\,{\scriptsize (+0.2)}}\\
            CSwin-B 
            & ${224}^2$     & 78M       & 15.0G     & 84.2\\
            \rowcolor{lightgray!50} \textbf{FLatten-CSwin-B} 
            & ${224}^2$     & 75M       & 15.0G     & \textbf{84.5\,{\scriptsize (+0.3)}}\\
            CSwin-B 
            & ${384}^2$     & 78M       & 47.0G     & 85.4\\
            \rowcolor{lightgray!50} \textbf{FLatten-CSwin-B} 
            & ${384}^2$     & 78M       & 46.4G     & \textbf{85.5\,{\scriptsize (+0.1)}}\\
            \bottomrule
        \end{tabular}}
    \vskip 0.15in
    \caption{Comparisons of focused linear attention with other vision transformer backbones on the ImageNet-1K classification task.}
    \label{tab:main_results}
\end{table}

\begin{table*}[h]\small
    \centering
    \setlength{\tabcolsep}{3.5mm}{
    \renewcommand\arraystretch{1.2}
    \begin{tabular}{c|c|c|c}
    \bottomrule
    \multirow{2}*{stage} & \multirow{2}*{output} & \multicolumn{2}{c}{FLatten-DeiT-T}\\
    \cline{3-4}
    & & \textbf{FLatten} & DeiT Block \\
    \hline
    \multirow{3}*[0.17in]{res1} & \multirow{3}*[0.17in]{$14\times 14$} & $\left[\!\!\! \begin{array}{c} {\rm \ win}  \ 14\!\times\! 14\\{\rm dim} \ 192 \\ {\rm head} \ 3\end{array} \!\!\! \right ] \!\!\times\! 12$ & None \\
    \toprule
    \end{tabular}}
    \vskip 0.05in
    \caption{Architectures of FLatten-DeiT models.}
    \label{tab:model_deit}
\end{table*}

\begin{table*}[h]\small
    \centering
    \setlength{\tabcolsep}{3.5mm}{
    \renewcommand\arraystretch{1.2}
    \begin{tabular}{c|c|c|c|c|c}
    \bottomrule
    \multirow{2}*{stage} & \multirow{2}*{output} & \multicolumn{2}{c|}{FLatten-PVT-M} & \multicolumn{2}{c|}{FLatten-PVT-L}\\
    \cline{3-6}
    & & \textbf{FLatten} & PVT Block & \textbf{FLatten} & PVT Block \\
    \hline
    \multirow{4}*{res1} & \multirow{4}*{$56\times 56$} & \multicolumn{4}{c}{Conv1×1, stride=4, 64, LN}\\
    \cline{3-6}
    && $\left[\!\!\! \begin{array}{c} {\rm \ win}  \ 56\!\times\! 56\\{\rm dim} \ 64 \\ {\rm head} \ 1\end{array} \!\!\! \right ] \!\!\times\! 2$ & None & $\left[\!\!\! \begin{array}{c} {\rm \ win}  \ 56\!\times\! 56\\{\rm dim} \ 64 \\ {\rm head} \ 1\end{array} \!\!\! \right ] \!\!\times\! 3$ & None \\
    \hline
    \multirow{4}*{res2} & \multirow{4}*{$28\times 28$} & \multicolumn{4}{c}{Conv1×1, stride=2, 128, LN}\\
    \cline{3-6}
    && $\left[\!\!\! \begin{array}{c} {\rm \ win}  \ 28\!\times\! 28\\{\rm dim} \ 128 \\ {\rm head} \ 2\end{array} \!\!\! \right ] \!\!\times\! 2$ & None & $\left[\!\!\! \begin{array}{c} {\rm \ win}  \ 28\!\times\! 28\\{\rm dim} \ 128 \\ {\rm head} \ 2\end{array} \!\!\! \right ] \!\!\times\! 3$ & None \\
    \hline
    \multirow{4}*{res3} & \multirow{4}*{$14\times 14$} & \multicolumn{4}{c}{Conv1×1, stride=2, 320, LN}\\
    \cline{3-6}
    & & $\left[\!\!\! \begin{array}{c} {\rm \ win}  \ 14\!\times\! 14\\{\rm dim} \ 320 \\ {\rm head} \ 5\end{array} \!\!\! \right ] \!\!\times\! 2$ & None & $\left[\!\!\! \begin{array}{c} {\rm \ win}  \ 14\!\times\! 14\\{\rm dim} \ 320 \\ {\rm head} \ 5\end{array} \!\!\! \right ] \!\!\times\! 6$ & None \\
    \hline
    \multirow{4}*{res4} & \multirow{4}*{$7\times 7$} & \multicolumn{4}{c}{Conv1×1, stride=2, 512, LN}\\
    \cline{3-6}
    & & $\left[\!\!\! \begin{array}{c} {\rm \ win}  \ 7\!\times\! 7\\{\rm dim} \ 512 \\ {\rm head} \ 8\end{array} \!\!\! \right ] \!\!\times\! 2$ & None & $\left[\!\!\! \begin{array}{c} {\rm \ win}  \ 7\!\times\! 7\\{\rm dim} \ 512 \\ {\rm head} \ 8\end{array} \!\!\! \right ] \!\!\times\! 3$ & None \\
    \toprule
    \end{tabular}}
    \vskip 0.05in
    \caption{Architectures of FLatten-PVT models (Part1).}
    \label{tab:model_pvt-1}
\end{table*}

\begin{table*}[h]\small
    \centering
    \setlength{\tabcolsep}{3.5mm}{
    \renewcommand\arraystretch{1.2}
    \begin{tabular}{c|c|c|c|c|c}
    \bottomrule
    \multirow{2}*{stage} & \multirow{2}*{output} & \multicolumn{2}{c|}{FLatten-PVT-M} & \multicolumn{2}{c|}{FLatten-PVT-L}\\
    \cline{3-6}
    & & \textbf{FLatten} & PVT Block & \textbf{FLatten} & PVT Block \\
    \hline
    \multirow{4}*{res1} & \multirow{4}*{$56\times 56$} & \multicolumn{4}{c}{Conv1×1, stride=4, 64, LN}\\
    \cline{3-6}
    && $\left[\!\!\! \begin{array}{c} {\rm \ win}  \ 56\!\times\! 56\\{\rm dim} \ 64 \\ {\rm head} \ 1\end{array} \!\!\! \right ] \!\!\times\! 3$ & None & $\left[\!\!\! \begin{array}{c} {\rm \ win}  \ 56\!\times\! 56\\{\rm dim} \ 64 \\ {\rm head} \ 1\end{array} \!\!\! \right ] \!\!\times\! 3$ & None \\
    \hline
    \multirow{4}*{res2} & \multirow{4}*{$28\times 28$} & \multicolumn{4}{c}{Conv1×1, stride=2, 128, LN}\\
    \cline{3-6}
    && $\left[\!\!\! \begin{array}{c} {\rm \ win}  \ 28\!\times\! 28\\{\rm dim} \ 128 \\ {\rm head} \ 2\end{array} \!\!\! \right ] \!\!\times\! 3$ & None & $\left[\!\!\! \begin{array}{c} {\rm \ win}  \ 28\!\times\! 28\\{\rm dim} \ 128 \\ {\rm head} \ 2\end{array} \!\!\! \right ] \!\!\times\! 8$ & None \\
    \hline
    \multirow{4}*{res3} & \multirow{4}*{$14\times 14$} & \multicolumn{4}{c}{Conv1×1, stride=2, 320, LN}\\
    \cline{3-6}
    & & $\left[\!\!\! \begin{array}{c} {\rm \ win}  \ 14\!\times\! 14\\{\rm dim} \ 320 \\ {\rm head} \ 5\end{array} \!\!\! \right ] \!\!\times\! 18$ & None & $\left[\!\!\! \begin{array}{c} {\rm \ win}  \ 14\!\times\! 14\\{\rm dim} \ 320 \\ {\rm head} \ 5\end{array} \!\!\! \right ] \!\!\times\! 27$ & None \\
    \hline
    \multirow{4}*{res4} & \multirow{4}*{$7\times 7$} & \multicolumn{4}{c}{Conv1×1, stride=2, 512, LN}\\
    \cline{3-6}
    & & $\left[\!\!\! \begin{array}{c} {\rm \ win}  \ 7\!\times\! 7\\{\rm dim} \ 512 \\ {\rm head} \ 8\end{array} \!\!\! \right ] \!\!\times\! 3$ & None & $\left[\!\!\! \begin{array}{c} {\rm \ win}  \ 7\!\times\! 7\\{\rm dim} \ 512 \\ {\rm head} \ 8\end{array} \!\!\! \right ] \!\!\times\! 3$ & None \\
    \toprule
    \end{tabular}}
    \vskip 0.05in
    \caption{Architectures of FLatten-PVT models (Part2).}
    \label{tab:model_pvt-2}
\end{table*}

\begin{table*}\small
    \centering
    \setlength{\tabcolsep}{1mm}{
    \renewcommand\arraystretch{1.15}
    \begin{tabular}{c|c|c|c|c|c|c|c}
    \bottomrule
    \multirow{2}*{stage} & \multirow{2}*{output} & \multicolumn{2}{c|}{FLatten-PVTv2-B0} & \multicolumn{2}{c|}{FLatten-PVTv2-B1} & \multicolumn{2}{c}{FLatten-PVTv2-B2}\\
    \cline{3-8}
    & & \textbf{FLatten} & PVTv2 Block & \textbf{FLatten} & PVTv2 Block& \textbf{FLatten} & PVTv2 Block\\
    \hline
    \multirow{4}*{res1} & \multirow{4}*{$56\times 56$} & \multicolumn{2}{c|}{Conv4×4, stride=4, 32, LN} & \multicolumn{4}{c}{Conv4×4, stride=4, 64, LN}\\
    \cline{3-8}
    && $\left[\!\!\! \begin{array}{c} {\rm \ win}  \ 56\!\times\! 56\\{\rm dim} \ 32 \\ {\rm head} \ 1\end{array} \!\!\! \right ] \!\!\times\! 2$ & None & $\left[\!\!\! \begin{array}{c} {\rm \ win}  \ 56\!\times\! 56\\{\rm dim} \ 64 \\ {\rm head} \ 1\end{array} \!\!\! \right ] \!\!\times\! 2$ & None & $\left[\!\!\! \begin{array}{c} {\rm \ win}  \ 56\!\times\! 56\\{\rm dim} \ 64 \\ {\rm head} \ 1\end{array} \!\!\! \right ] \!\!\times\! 3$ & None\\
    \hline
    \multirow{4}*{res2} & \multirow{4}*{$28\times 28$} & \multicolumn{2}{c|}{Conv1×1, stride=2, 64, LN} & \multicolumn{4}{c}{Conv1×1, stride=2, 128, LN}\\
    \cline{3-8}
    && $\left[\!\!\! \begin{array}{c} {\rm \ win}  \ 28\!\times\! 28\\{\rm dim} \ 64 \\ {\rm head} \ 2\end{array} \!\!\! \right ] \!\!\times\! 2$ & None & $\left[\!\!\! \begin{array}{c} {\rm \ win}  \ 28\!\times\! 28\\{\rm dim} \ 128 \\ {\rm head} \ 2\end{array} \!\!\! \right ] \!\!\times\! 2$ & None & $\left[\!\!\! \begin{array}{c} {\rm \ win}  \ 28\!\times\! 28\\{\rm dim} \ 128 \\ {\rm head} \ 2\end{array} \!\!\! \right ] \!\!\times\! 3$ & None\\
    \hline
    \multirow{4}*{res3} & \multirow{4}*{$14\times 14$}  & \multicolumn{2}{c|}{Conv2×2, stride=2, 160, LN} & \multicolumn{4}{c}{Conv2×2, stride=2, 320, LN}\\
    \cline{3-8}
    & & $\left[\!\!\! \begin{array}{c} {\rm \ win}  \ 14\!\times\! 14\\{\rm dim} \ 160 \\ {\rm head} \ 5\end{array} \!\!\! \right ] \!\!\times\! 2$ & None & $\left[\!\!\! \begin{array}{c} {\rm \ win}  \ 14\!\times\! 14\\{\rm dim} \ 320 \\ {\rm head} \ 5\end{array} \!\!\! \right ] \!\!\times\! 2$ & None & $\left[\!\!\! \begin{array}{c} {\rm \ win}  \ 14\!\times\! 14\\{\rm dim} \ 320 \\ {\rm head} \ 5\end{array} \!\!\! \right ] \!\!\times\! 6$ & None\\
    \hline
    \multirow{4}*{res4} & \multirow{4}*{$7\times 7$}  & \multicolumn{2}{c|}{Conv2×2, stride=2, 256, LN} & \multicolumn{4}{c}{Conv2×2, stride=2, 512, LN}\\
    \cline{3-8}
    & & $\left[\!\!\! \begin{array}{c} {\rm \ win}  \ 7\!\times\! 7\\{\rm dim} \ 512 \\ {\rm head} \ 8\end{array} \!\!\! \right ] \!\!\times\! 2$ & None & $\left[\!\!\! \begin{array}{c} {\rm \ win}  \ 7\!\times\! 7\\{\rm dim} \ 512 \\ {\rm head} \ 8\end{array} \!\!\! \right ] \!\!\times\! 2$ & None & $\left[\!\!\! \begin{array}{c} {\rm \ win}  \ 7\!\times\! 7\\{\rm dim} \ 512 \\ {\rm head} \ 8\end{array} \!\!\! \right ] \!\!\times\! 3$ & None\\
    \toprule
    \end{tabular}}
    \caption{Architectures of FLatten-PVTv2 models (Part1).}
    \label{tab:model_pvtv2-1}
\end{table*}

\begin{table*}\small
    \centering
    \setlength{\tabcolsep}{3.5mm}{
    \renewcommand\arraystretch{1.15}
    \begin{tabular}{c|c|c|c|c|c}
    \bottomrule
    \multirow{2}*{stage} & \multirow{2}*{output} & \multicolumn{2}{c|}{FLatten-PVTv2-B3} & \multicolumn{2}{c|}{FLatten-PVTv2-B4}\\
    \cline{3-6}
    & & \textbf{FLatten} & PVTv2 Block & \textbf{FLatten} & PVTv2 Block\\
    \hline
    \multirow{4}*{res1} & \multirow{4}*{$56\times 56$} & \multicolumn{4}{c}{Conv4×4, stride=4, 64, LN}\\
    \cline{3-6}
    && $\left[\!\!\! \begin{array}{c} {\rm \ win}  \ 56\!\times\! 56\\{\rm dim} \ 64 \\ {\rm head} \ 1\end{array} \!\!\! \right ] \!\!\times\! 3$ & None & $\left[\!\!\! \begin{array}{c} {\rm \ win}  \ 56\!\times\! 56\\{\rm dim} \ 64 \\ {\rm head} \ 1\end{array} \!\!\! \right ] \!\!\times\! 3$ & None\\
    \hline
    \multirow{4}*{res2} & \multirow{4}*{$28\times 28$} & \multicolumn{4}{c}{Conv2×2, stride=2, 128, LN}\\
    \cline{3-6}
    && $\left[\!\!\! \begin{array}{c} {\rm \ win}  \ 28\!\times\! 28\\{\rm dim} \ 128 \\ {\rm head} \ 2\end{array} \!\!\! \right ] \!\!\times\! 3$ & None & $\left[\!\!\! \begin{array}{c} {\rm \ win}  \ 28\!\times\! 28\\{\rm dim} \ 128 \\ {\rm head} \ 2\end{array} \!\!\! \right ] \!\!\times\! 8$ & None\\
    \hline
    \multirow{4}*{res3} & \multirow{4}*{$14\times 14$}  & \multicolumn{4}{c}{Conv2×2, stride=2, 320, LN}\\
    \cline{3-6}
    && $\left[\!\!\! \begin{array}{c} {\rm \ win}  \ 14\!\times\! 14\\{\rm dim} \ 320 \\ {\rm head} \ 5\end{array} \!\!\! \right ] \!\!\times\! 18$ & None & $\left[\!\!\! \begin{array}{c} {\rm \ win}  \ 14\!\times\! 14\\{\rm dim} \ 320 \\ {\rm head} \ 5\end{array} \!\!\! \right ] \!\!\times\! 27$ & None\\
    \hline
    \multirow{4}*{res4} & \multirow{4}*{$7\times 7$}  & \multicolumn{4}{c}{Conv1×1, stride=2, 512, LN}\\
    \cline{3-6}
    & & $\left[\!\!\! \begin{array}{c} {\rm \ win}  \ 7\!\times\! 7\\{\rm dim} \ 512 \\ {\rm head} \ 8\end{array} \!\!\! \right ] \!\!\times\! 3$ & None & $\left[\!\!\! \begin{array}{c} {\rm \ win}  \ 7\!\times\! 7\\{\rm dim} \ 512 \\ {\rm head} \ 8\end{array} \!\!\! \right ] \!\!\times\! 3$ & None\\
    \toprule
    \end{tabular}}
    \caption{Architectures of FLatten-PVTv2 models (Part2).}
    \label{tab:model_pvtv2-2}
\end{table*}

\begin{table*}\small
    \centering
    \setlength{\tabcolsep}{0.5mm}{
    \renewcommand\arraystretch{1.15}
    \begin{tabular}{c|c|c|c|c|c|c|c}
    \bottomrule
    \multirow{2}*{stage} & \multirow{2}*{output} & \multicolumn{2}{c|}{FLatten-Swin-T} & \multicolumn{2}{c|}{FLatten-Swin-S} & \multicolumn{2}{c}{FLatten-Swin-B}\\
    \cline{3-8}
    & & \textbf{FLatten} & Swin Block & \textbf{FLatten} & Swin Block& \textbf{FLatten} & Swin Block\\
    \hline
    \multirow{4}*{res1} & \multirow{4}*{$56\times 56$} & \multicolumn{2}{c|}{concat $4\times 4$, 96, LN} & \multicolumn{2}{c|}{concat $4\times 4$, 96, LN} & \multicolumn{2}{c}{concat $4\times 4$, 128, LN}\\
    \cline{3-8}
    && $\left[\!\!\! \begin{array}{c} {\rm \ win}  \ 56\!\times\! 56\\{\rm dim} \ 96 \\ {\rm head} \ 3\end{array} \!\!\! \right ] \!\!\times\! 2$ & None & $\left[\!\!\! \begin{array}{c} {\rm \ win}  \ 56\!\times\! 56\\{\rm dim} \ 96 \\ {\rm head} \ 3\end{array} \!\!\! \right ] \!\!\times\! 2$ & None & $\left[\!\!\! \begin{array}{c} {\rm \ win}  \ 56\!\times\! 56\\{\rm dim} \ 128 \\ {\rm head} \ 3\end{array} \!\!\! \right ] \!\!\times\! 2$ & None\\
    \hline
    \multirow{4}*{res2} & \multirow{4}*{$28\times 28$} & \multicolumn{2}{c|}{concat $4\times 4$, 192, LN} & \multicolumn{2}{c|}{concat $4\times 4$, 192, LN} & \multicolumn{2}{c}{concat $4\times 4$, 256, LN}\\
    \cline{3-8}
    && $\left[\!\!\! \begin{array}{c} {\rm \ win}  \ 28\!\times\! 28\\{\rm dim} \ 192 \\ {\rm head} \ 6\end{array} \!\!\! \right ] \!\!\times\! 2$ & None & $\left[\!\!\! \begin{array}{c} {\rm \ win}  \ 28\!\times\! 28\\{\rm dim} \ 192 \\ {\rm head} \ 6\end{array} \!\!\! \right ] \!\!\times\! 2$ & None & $\left[\!\!\! \begin{array}{c} {\rm \ win}  \ 28\!\times\! 28\\{\rm dim} \ 256 \\ {\rm head} \ 6\end{array} \!\!\! \right ] \!\!\times\! 2$ & None\\
    \hline
    \multirow{4}*{res3} & \multirow{4}*{$14\times 14$} & \multicolumn{2}{c|}{concat $4\times 4$, 384, LN} & \multicolumn{2}{c|}{concat $4\times 4$, 384, LN} & \multicolumn{2}{c}{concat $4\times 4$, 512, LN}\\
    \cline{3-8}
    && None & $\left[\!\!\! \begin{array}{c} {\rm \ win}  \ 7\!\times\! 7\\{\rm dim} \ 384 \\ {\rm head} \ 12\end{array} \!\!\! \right ] \!\!\times\! 6$ & None & $\left[\!\!\! \begin{array}{c} {\rm \ win}  \ 7\!\times\! 7\\{\rm dim} \ 384 \\ {\rm head} \ 12\end{array} \!\!\! \right ] \!\!\times\! 18$ & None & $\left[\!\!\! \begin{array}{c} {\rm \ win}  \ 7\!\times\! 7\\{\rm dim} \ 512 \\ {\rm head} \ 12\end{array} \!\!\! \right ] \!\!\times\! 18$\\
    \hline
    \multirow{4}*{res4} & \multirow{4}*{$7\times 7$} & \multicolumn{2}{c|}{concat $4\times 4$, 768, LN} & \multicolumn{2}{c|}{concat $4\times 4$, 768, LN} & \multicolumn{2}{c}{concat $4\times 4$, 1024, LN}\\
    \cline{3-8}
    & & None& $\left[\!\!\! \begin{array}{c} {\rm \ win}  \ 7\!\times\! 7\\{\rm dim} \ 768 \\ {\rm head} \ 24\end{array} \!\!\! \right ] \!\!\times\! 2$ & None & $\left[\!\!\! \begin{array}{c} {\rm \ win}  \ 7\!\times\! 7\\{\rm dim} \ 768 \\ {\rm head} \ 24\end{array} \!\!\! \right ] \!\!\times\! 2$ & None & $\left[\!\!\! \begin{array}{c} {\rm \ win}  \ 7\!\times\! 7\\{\rm dim} \ 1024 \\ {\rm head} \ 24\end{array} \!\!\! \right ] \!\!\times\! 2$\\
    \toprule
    \end{tabular}}
    \caption{Architectures of FLatten-Swin models.}
    \label{tab:model_swin}
\end{table*}

\begin{table*}\small
    \centering
    \setlength{\tabcolsep}{1mm}{
    \renewcommand\arraystretch{1.15}
    \begin{tabular}{c|c|c|c|c|c|c|c}
    \bottomrule
    \multirow{2}*{stage} & \multirow{2}*{output} & \multicolumn{2}{c|}{FLatten-CSwin-T} & \multicolumn{2}{c|}{FLatten-CSwin-S} & \multicolumn{2}{c}{FLatten-CSwin-B}\\
    \cline{3-8}
    & & \textbf{FLatten} & CSwin Block & \textbf{FLatten} & CSwin Block& \textbf{FLatten} & CSwin Block\\
    \hline
    \multirow{4}*{res1} & \multirow{4}*{$56\times 56$} & \multicolumn{4}{c|}{Conv7×7, stride=4, 64, LN} & \multicolumn{2}{c}{Conv7×7, stride=4, 96, LN} \\
    \cline{3-8}
    && $\left[\!\!\! \begin{array}{c} {\rm \ win}  \ 3\!\times\! 3\\{\rm dim} \ 64 \\ {\rm head} \ 2\end{array} \!\!\! \right ] \!\!\times\! 2$ & None & $\left[\!\!\! \begin{array}{c} {\rm \ win}  \ 3\!\times\! 3\\{\rm dim} \ 64 \\ {\rm head} \ 2\end{array} \!\!\! \right ] \!\!\times\! 2$ & None & $\left[\!\!\! \begin{array}{c} {\rm \ win}  \ 3\!\times\! 3\\{\rm dim} \ 96 \\ {\rm head} \ 4\end{array} \!\!\! \right ] \!\!\times\! 3$ & None\\
    \hline
    \multirow{4}*{res2} & \multirow{4}*{$28\times 28$} & \multicolumn{4}{c|}{Conv7×7, stride=4, 128, LN} & \multicolumn{2}{c}{Conv7×7, stride=4, 192, LN} \\
    \cline{3-8}
    && $\left[\!\!\! \begin{array}{c} {\rm \ win}  \ 3\!\times\! 3\\{\rm dim} \ 128 \\ {\rm head} \ 4\end{array} \!\!\! \right ] \!\!\times\! 4$ & None & $\left[\!\!\! \begin{array}{c} {\rm \ win}  \ 3\!\times\! 3\\{\rm dim} \ 128 \\ {\rm head} \ 4\end{array} \!\!\! \right ] \!\!\times\! 4$ & None & $\left[\!\!\! \begin{array}{c} {\rm \ win}  \ 3\!\times\! 3\\{\rm dim} \ 192 \\ {\rm head} \ 8\end{array} \!\!\! \right ] \!\!\times\! 6$ & None\\
    \hline
    \multirow{4}*{res3} & \multirow{4}*{$14\times 14$} & \multicolumn{4}{c|}{Conv7×7, stride=4, 256, LN} & \multicolumn{2}{c}{Conv7×7, stride=384, LN} \\
    \cline{3-8}
    & & None & $\left[\!\!\! \begin{array}{c} {\rm \ win}  \ 3\!\times\! 3\\{\rm dim} \ 256 \\ {\rm head} \ 8\end{array} \!\!\! \right ] \!\!\times\! 18$ & None &  $\left[\!\!\! \begin{array}{c} {\rm \ win}  \ 3\!\times\! 3\\{\rm dim} \ 256 \\ {\rm head} \ 8\end{array} \!\!\! \right ] \!\!\times\! 32$ & None & $\left[\!\!\! \begin{array}{c} {\rm \ win}  \ 3\!\times\! 3\\{\rm dim} \ 384 \\ {\rm head} \ 16\end{array} \!\!\! \right ] \!\!\times\! 29$\\
    \hline
    \multirow{4}*{res4} & \multirow{4}*{$7\times 7$} & \multicolumn{4}{c|}{Conv7×7, stride=4, 512, LN} & \multicolumn{2}{c}{Conv7×7, stride=4, 768, LN} \\
    \cline{3-8}
    & & None& $\left[\!\!\! \begin{array}{c} {\rm \ win}  \ 7\!\times\! 7\\{\rm dim} \ 512 \\ {\rm head} \ 16\end{array} \!\!\! \right ] \!\!\times\! 1$ & None & $\left[\!\!\! \begin{array}{c} {\rm \ win}  \ 7\!\times\! 7\\{\rm dim} \ 512 \\ {\rm head} \ 16\end{array} \!\!\! \right ] \!\!\times\! 2$ & None & $\left[\!\!\! \begin{array}{c} {\rm \ win}  \ 7\!\times\! 7\\{\rm dim} \ 768 \\ {\rm head} \ 32\end{array} \!\!\! \right ] \!\!\times\! 2$\\
    \toprule
    \end{tabular}}
    \caption{Architectures of FLatten-CSwin models.}
    \label{tab:model_cswin}
\end{table*}

\section*{D. Model Architectures}

We summarize the architectures of five Transformer models adopted in the main paper, including DeiT~\cite{deit}, PVT~\cite{pvt}, PVTv2~\cite{pvtv2}, Swin Transformer~\cite{swin}, CSwin Transformer~\cite{cswin} in Tab.\ref{tab:model_deit}-\ref{tab:model_cswin}. In practice, we substitute the original self-attention blocks at all stages of the DeiT, PVT and PVTv2 with the focused linear attention block, but only adopt our module at early stages of Swin and CSwin. The model structure (width and depth) are kept unchanged, except for CSwin-T and CSwin-B, where we increase the depth of the first and second stages and correspondingly reduce the depth of the third stage to better reflect our module's advantage of enlarged receptive field.

\end{document}